\documentclass[10pt,twocolumn,letterpaper]{article}

\usepackage{cvpr}
\usepackage{times}
\usepackage{epsfig}
\usepackage{graphicx}
\usepackage{amsmath}
\usepackage{amssymb}
\usepackage{booktabs}
\usepackage{xcolor}
\usepackage{multirow}
\usepackage{float}
\usepackage{amsfonts}
\usepackage{comment}
\usepackage{url}
\usepackage[printonlyused]{acronym}
\usepackage[pagebackref=true,breaklinks=true,letterpaper=true,colorlinks,bookmarks=false]{hyperref}
\pdfoutput=1
\acrodef{CNN}{Convolutional Neural Network}
\acrodef{CNNs}{Convolutional Neural Networks}
\acrodef{ROC}{Receiving Operating Characteristic}
\acrodef{AMD}{Age-related Macular Degeneration}
\acrodef{DoG}{Difference Of Gaussian}
\acrodef{NMS}{Non-Max-Supression}
\acrodef{SIFT}{Scale-Invariant Feature Transform}
\acrodef{RanSaC}{Random Sampling Consensus}
\acrodef{RL}{Reinforcement Learning}
\acrodef{NNDR}{Nearest Neighbor Distance Ratio}
\acrodef{DR}{Diabetic Retinopathy}
\acrodef{SURF}{Speeded-Up Robust Features}
\acrodef{FIRE}{Fundus Image Registration}
\acrodef{ORB}{Oriented Fast and Rotated Brief}
\acrodef{FREAK}{Fast Retina Keypoint}
\acrodef{BRISK}{Binary Robust Invariant Scalable Keypoints}
\acrodef{SfM}{Structure from Motion}
\acrodef{LIFT}{Learned Invariant Feature Transform  }
\acrodef{GLAMpoints}{Greedily Learned Accurate Match Points}
\acrodef{SLAM}{Simultaneous Localization and Mapping}
\acrodef{ReLU}{Rectified Linear Unit}
\acrodef{UNet}{U-Net}
\acrodef{LF-NET}{Local Feature Network}
\newcommand{\parsection}[1]{\noindent\textbf{#1:} }

\cvprfinalcopy

\ifcvprfinal\fi
\begin{document}

\title{GLU-Net: Global-Local Universal Network for\\Dense Flow and Correspondences}

\author{Prune Truong \quad Martin Danelljan \quad Radu Timofte \\
Computer Vision Lab, D-ITET, ETH Z\"urich, Switzerland \\
{\tt\small \{truongp,martin.danelljan,radu.timofte\}@ethz.ch}
}

\maketitle
\begin{abstract}
    Establishing dense correspondences between a pair of images is an important and general problem, covering geometric matching, optical flow and semantic correspondences. While these applications share fundamental challenges, such as large displacements, pixel-accuracy, and appearance changes, they are currently addressed with specialized network architectures, designed for only one particular task. This severely limits the generalization capabilities of such networks to new scenarios, where \eg robustness to larger displacements or higher accuracy is required.

In this work, we propose a universal network architecture that is directly applicable to all the aforementioned dense correspondence problems. We achieve both high accuracy and robustness to large displacements by investigating the combined use of global and local correlation layers. We further propose an adaptive resolution strategy, allowing our network to operate on virtually any input image resolution. 
The proposed \textbf{GLU-Net} achieves state-of-the-art performance for geometric and semantic matching as well as optical flow, when using the \emph{same} network and weights. Code and trained models are available at \mbox{\url{https://github.com/PruneTruong/GLU-Net}}.

\end{abstract}

\section{Introduction}

Finding pixel-to-pixel correspondences between images continues to be a fundamental problem in Computer Vision~\cite{Forsyth,Hartley}. This is due to its many important applications, including visual localization~\cite{johannes2017, Taira2018}, 3D-reconstruction~\cite{AgarwalFSSCSS11}, structure-from-motion~\cite{SchonbergerF16}, image manipulation~\cite{HaCohenSGL11, LiuYT11}, action recognition~\cite{SimonyanZ14} and autonomous driving~\cite{JanaiGBG17}. Due to the astonishing performance achieved by the recent developments in deep learning, end-to-end trainable \ac{CNNs} are now applied for this task in all the aforementioned domains.

\begin{figure}[t]
\centering
\vspace{-2mm}\includegraphics[width=0.48\textwidth]{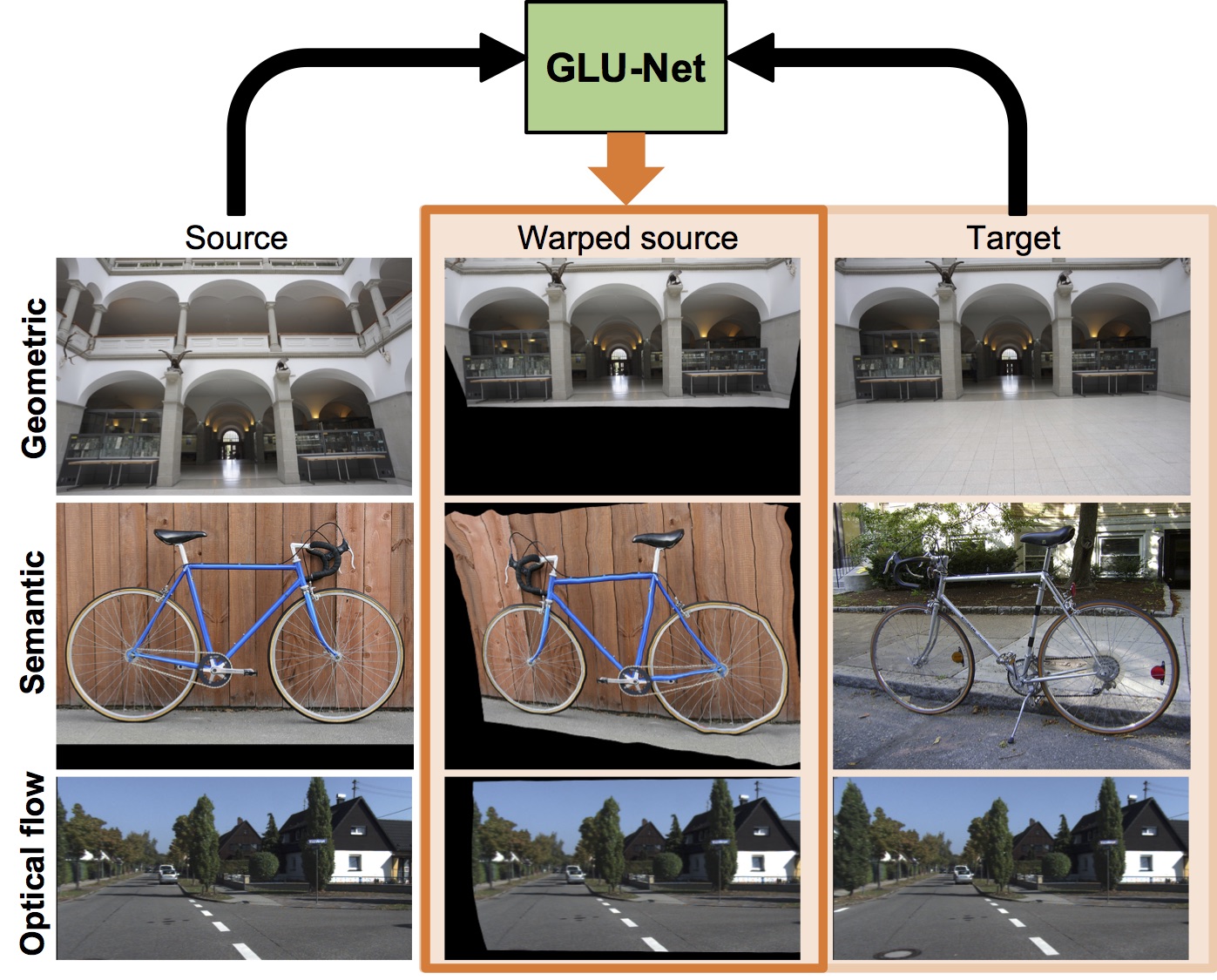}
\caption{Our GLU-Net estimates dense correspondences between a source (left) and a target (right) image. The estimated correspondences are here used to warp (center) the source image. The warped result (center) accurately matches the target image (right). The \emph{same} network and weights are applied for Geometric matching, Semantic matching and Optical flow tasks.}\vspace{-4.5mm}
\label{intro}
\end{figure}

The general problem of estimating correspondences between pairs of images can be divided into several different tasks, depending on the origin of the images.  In the \emph{geometric matching} task~\cite{Hartley}, the images constitute different views of the same scene, taken by a single or multiple cameras. The images may be taken from radically different viewpoints, leading to large displacements and appearance transformations between the frames.
On the other hand, \emph{optical flow}~\cite{Baker2011,Horn1981} aims to estimate accurate pixel-wise displacements between two consecutive frames of a sequence or video. In the \emph{semantic matching} problem~\cite{Ham2016,LiuYT11} (also referred as semantic flow),  the task is instead to find semantically meaningful correspondences between different instances of the same scene category or object, such as `car' or `horse'.
Current methods generally address \emph{one} of these tasks, using specialized architectures that generalize poorly to related correspondence problems. In this work, we therefore set out to design a universal architecture that jointly addresses all aforementioned tasks. 

One key architectural aspect shared by a variety of correspondence networks is the reliance on \emph{correlation layers}, computing local similarities between deep features extracted from the two images. These provide strong cues when establishing correspondences.
Optical flow methods typically employ \emph{local} correlation layers~\cite{Dosovitskiy2015, Hui2018, Hui2019, Ilg2017a, Sun2018, Sun2019}, evaluating similarities in a local neighborhood around an image coordinate. While suitable for small displacements, they are unable to capture large viewpoints changes. On the contrary, geometric and semantic matching architectures utilize \emph{global} correlations~\cite{Kim2019, Melekhov2019, Rocco2017a, Rocco2018a, Rocco2018b}, where similarities are evaluated between all pairs of locations in the dense feature maps. While capable of handling long-range matches, global correlation layers are computationally unfeasible at high resolutions. Moreover, they constrain the input image size to a pre-determined resolution, which severely hampers accuracy for high-resolution images. 

\newcommand{\bp}[1]{\textbf{#1}}

\parsection{Contributions}
In this paper, we propose GLU-Net, a Global-Local Universal Network for estimating dense correspondences. 
Our architecture is robust to large viewpoint changes and appearance transformations, while capable of estimating small displacements with high accuracy. 
The main contributions of this work are: 
\bp{(i)} We introduce a single unified architecture, applicable to geometric matching, semantic matching and optical flow. 
\bp{(ii)} Our network carefully integrates global and local correlation layers to handle both large and small displacements. 
\bp{(iii)} To circumvent the fixed input resolution imposed by the global cost volume, we propose an adaptive resolution strategy that enables our network to take \emph{any} image resolution as input, crucial for high-accuracy displacements. 
\bp{(iv)} We train our network in a self-supervised manner, relying on synthetic warps of real images, thus requiring no annotated ground-truth flow. 

We perform comprehensive experiments on the three aforementioned tasks, providing detailed analysis of our approach and thorough comparisons with recent state-of-the-art. 
Our approach outperforms previous methods for dense geometric correspondences on the HPatches~\cite{Lenc} and ETH3D~\cite{ETH3d} datasets, while setting a new state-of-the-art for semantic correspondences on the TSS~\cite{Taniai2016} dataset. Moreover, our network, without any retraining or fine-tuning, generalizes to optical flow by providing highly competitive results on the KITTI~\cite{Geiger2013} dataset. Both training code and models are available at~\cite{glu-net}.

\section{Related work}

Finding correspondences between a pair of images is a classical computer vision problem, uniting optical flow, geometric correspondences and semantic matching. This problem dates back several decades~\cite{Horn1981}, with most classical techniques relying on hand crafted~\cite{Alahi2012,Alcantarilla2012,Bay2006,harris,Leutenegger2011,Lowe2004,Rublee2011} or trained~\cite{superpoint, OnoTFY18,GLAMpoint} feature detectors/descriptors, or variational formulations~\cite{Baker2011,Horn1981,LiuYT11}. In recent years, \ac{CNNs} have revolutionised most areas within vision, including different aspects of the image correspondence problem. Here, we focus on \ac{CNN}-based methods for generating \emph{dense} correspondences or flow fields, as these are most related to our work.

\parsection{Optical Flow}
Dosovitskiy~\etal~\cite{Dosovitskiy2015} constructed the first trainable \ac{CNN} for optical flow estimation, FlowNet, based on a U-Net denoising autoencoder architecture~\cite{Vincent2010} and trained it on the large synthetic FlyingChairs dataset. Ilg~\etal~\cite{Ilg2017a} stacked several basic FlowNet models into a large one, called FlowNet2, which performed on par with classical state-of-the-art methods on the Sintel benchmark~\cite{Butler2012}. Subsequently, Ranjan and Black~\cite{Ranjan2017} introduced SpyNet, a compact spatial image pyramid network. 

Recent notable contributions to end-to-end trainable optical flow include PWC-Net~\cite{Sun2018, Sun2019} and LiteFlowNet~\cite{Hui2018}, followed by LiteFlowNet2~\cite{Hui2019}. They employ multiple constrained correlation layers operating on a feature pyramid, where the features at each level are warped by the current flow estimate, yielding more compact and effective networks. 
Nevertheless, while these networks excel at small to medium displacements with small appearance changes, they perform poorly on strong geometric transformations or when the visual appearance is significantly different. 

\parsection{Geometric Correspondence}
Unlike optical flow, geometric correspondence estimation focuses on \emph{large} geometric displacements, which can cause significant appearance distortions between the frames.
Motivated by recent advancements in optical flow architectures, Melekhov~\etal~\cite{Melekhov2019} introduced DGC-Net, a coarse-to-fine \ac{CNN}-based framework that generates dense 2D correspondences between image pairs. It relies on a global cost volume constructed at the coarsest resolution.
However, the input size is constrained to a fixed resolution ($240 \times 240$), severely limiting its performance on higher resolution images.
Rocco~\etal~\cite{Rocco2018b} aim at increasing the performance of the global correlation layer by proposing an end-to-end trainable neighborhood consensus network, NC-Net, to filter out ambiguous matches and keep only the locally and cyclically consistent ones. 
Furthermore, Laskar~\etal~\cite{Laskar2019} utilize a modified version of DGC-Net, focusing on image retrieval.

\parsection{Semantic Correspondence}
Unlike optical flow or geometric matching, semantic correspondence poses additional challenges due to intra-class appearance and shape variations among different instances from the same object or scene category. 
Rocco~\etal~\cite{Rocco2017a, Rocco2018a} proposed the CNNGeo matching architecture, predicting globally parametrized affine and thin plate spline transformations between image pairs. 
Other approaches aim to predict richer geometric deformations \cite{Choy2016,Kim2018, KimMHLS19, Rocco2018b} using \eg Spatial Transformer Networks \cite{Jaderberg2015}.
Recently, Jeon~\etal~\cite{Jeon} introduced PARN, a pyramidal model where dense affine transformation fields are progressively estimated in a coarse-to-fine manner. 
SAM-Net~\cite{Kim2019} obtains better results by jointly learning semantic correspondence and attribute transfer. 
Huang~\etal~\cite{DCCNet} proposed DCCNet, which fuses correlation maps derived from local features and a context-aware semantic feature representation. 

\section{Method}
\label{sec:method}

We address the problem of finding pixel-wise correspondences between a pair of images $I_{\text{s}} \in \mathbb{R}^{H \times W \times 3}$ and $I_{\text{t}} \in \mathbb{R}^{H \times W \times 3}$. In this work, we put no particular assumptions on the origin of the image pair itself. It may correspond to two different views of the same scene, two consecutive frames in a video, or two images with similar semantic content. Our goal is to estimate a dense displacement field, often referred to as flow, $\mathbf{w} \in \mathbb{R}^{H \times W \times 2}$ that warps image $I_\text{s}$ towards $I_\text{t}$ such that,
\begin{equation}
   I_\text{t}(\mathbf{x}) \approx I_\text{s}( \mathbf{x} + \mathbf{w}(\mathbf{x}))\,.
\end{equation}
The flow $\mathbf{w}$ represents the pixel-wise 2D motion vectors in the target image coordinate system. It is directly related to the pixel correspondence map $\mathbf{m}(\mathbf{x}) = \mathbf{x} + \mathbf{w}(\mathbf{x})$, which directly maps an image coordinate $\mathbf{x}$ in the target image to its corresponding position in the source image.  

In this work, we design an architecture capable of robustly finding both long-range correspondences and accurate estimation of pixel-wise displacements. We thereby achieve a universal network for predicting dense flow fields, applicable to geometric matching, semantic correspondences and optical flow. The overall architecture follows a CNN feature-based coarse-to-fine strategy, which has proved widely successful for specific tasks~\cite{Hui2018, Jeon,  Kim2019, Melekhov2019, Sun2018}. However, contrary to previous works, our architecture combines global and local correlation layers, as discussed in Section~\ref{subsec:loc-global} and~\ref{subsec:Mixed Global-Local}, to benefit from their complementary properties. We further circumvent the input resolution restriction imposed by the global correlation layer by introducing an adaptive resolution strategy in Section~\ref{subsec:adaptive_reso}. It is based on a two-stream feature pyramid, which allows dense correspondence prediction for \emph{any} input resolution image. Our final architecture is detailed in Section~\ref{subsec:architecture-det} and the training procedure explained in Section~\ref{subsec:training}.

\begin{figure*}[t]
\centering
\includegraphics[width=0.999\textwidth]{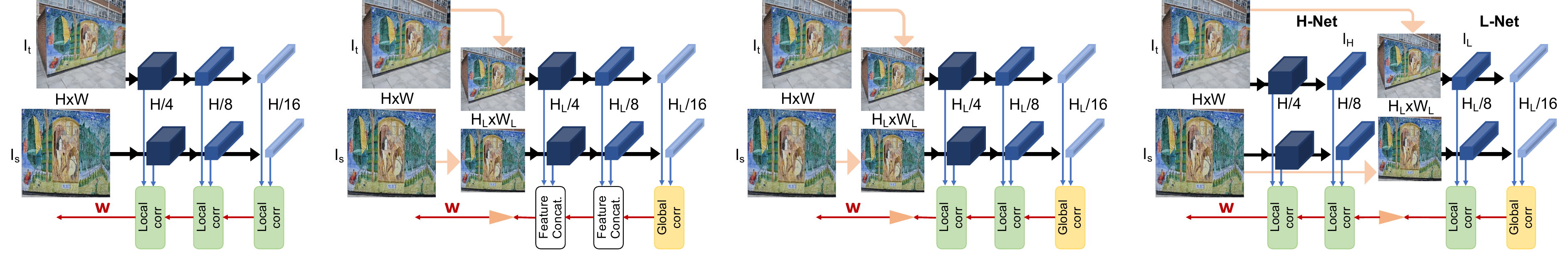}
\begin{tabular}{c}
\small{
    \hspace{0.3cm} (a) Local-Net \hspace{2.3cm} (b) Global-Net \hspace{2.3cm} (c) GLOCAL-Net  \hspace{2.3cm} (d) GLU-Net (Ours) \hspace{2cm}}
\end{tabular}
\caption{Schematic representation of different architectures for dense flow field estimation $\mathbf{w}$. Local-Net (a) and Global-Net (b) employ \emph{only} local and global correlation layers, respectively. Our GLOCAL-Net (c) combines both to effectively handle short and long-range displacements. GLU-Net (d) additionally employs our adaptive resolution strategy, thus capable of processing high-resolution images. 
}\vspace{-4mm}
\label{arch}
\end{figure*}

\subsection{Local and Global Correlations}
\label{subsec:loc-global}

Current state-of-the-art architectures~\cite{DCCNet,Hui2018,Jeon, Melekhov2019,Sun2018} for estimating image correspondences or optical flow rely on measuring local similarities between the source and target images. This is performed in a deep feature space, which provides a discriminative embedding with desirable invariances. 
The result, generally referred to as a correlation or cost volume, provides an extremely powerful cue when deriving the final correspondence or flow estimate. 
The correlation can be performed in a local or global manner.

\parsection{Local correlation} In a local correlation layer, the feature similarity is only evaluated in the neighborhood of the target image coordinate, specified by a search radius $R$. Formally, the correlation $c^l$ between the target $F_\text{t}^l \in \mathbb{R}^{H_{l} \times W_{l} \times d_l}$ and source $F_\text{s}^l \in \mathbb{R}^{H_{l} \times W_{l} \times d_{l}}$ feature maps is defined as,
\begin{equation}
\label{eq:local-corr}
c^l(\mathbf{x}, \mathbf{d}) = F_\text{t}^l(\mathbf{x})^{T}  F_\text{s}^l(\mathbf{x} + \mathbf{d}) \,, \quad \|\mathbf{d}\|_\infty \leq R \,,
\end{equation}
where $\mathbf{x} \in \mathbb{Z}^2$ is a coordinate in the target feature map and $\mathbf{d} \in \mathbb{Z}^2$ is the displacement from this location. The displacement is constrained to $\|\mathbf{d}\|_\infty \leq R$, \ie the maximum motion in any direction is $R$. We let $l$ denote the level in the feature pyramid. 
While most naturally thought of as a 4-dimensional tensor, the two displacement dimensions are usually vectorized into one to simplify further processing in the CNN. The resulting 3D correlation volume $c^l$ thus has a dimensionality of $H_l \times W_l \times (2R+1)^2$. 

\parsection{Global correlation} A global correlation layer evaluates the pairwise similarities between all locations in the target and source feature maps. The correlation volume $C^l \in \mathbb{R}^{H_l \times W_l \times H_l \times W_l}$ contains at each target image location $\mathbf{x}\in \mathbb{Z}^2$ the scalar products between corresponding feature vector $F^l_\text{t}(\mathbf{x})$ and the vectors $F^l_\text{s}(\mathbf{x'}) \in \mathbb{R}^{d}$ extracted from all source feature map coordinates $\mathbf{x'}$,
\begin{equation}
\label{eq:global-corr}
C^l(\mathbf{x}, \mathbf{x'})=F^l_\text{t}(\mathbf{x})^{T} F^l_\text{s}\left(\mathbf{x'}\right) \,.
\end{equation}
As for the local cost volume, we vectorize the source dimensions, leading to a 3D tensor of size $H_l \times W_l \times(H_l W_l)$.

\parsection{Comparison} Local and global correlation layers have a few key contrary properties and behaviors. Local correlations are popularly employed in architectures designed for optical flow~\cite{Dosovitskiy2015, Hui2018, Sun2018}, where the displacements are generally small. Thanks to their restricted search region, local correlation layers can be applied for high-resolution feature maps, which allows accurate estimation of small displacements. On the other hand, a local correlation based architecture is limited to a certain maximum range of displacements. 
Conversely, a global correlation based architecture does not suffer from this limitation, encapsulating arbitrary long-range displacements. 

The major disadvantage of the global cost volume is that its dimensionality scales with the size of the feature map $H_l \times W_l$. Therefore, due to the quadratic $\mathcal{O}((H_l W_l)^2)$ scaling in computation and memory, global cost volumes are only suitable at coarse resolutions. Moreover, post-processing layers implemented with 2D convolutions expect a fixed channel dimensionality. Since the  channel dimension $H_l W_l$ of the cost volume depends on its spatial dimensions $H_l \times W_l$, this effectively constrains the network input resolution to a fixed pre-determined value, referred to as $H_L \times W_L$. 
The network can thus not leverage the more detailed structure in high-resolution images and lacks precision, since the images require down-scaling to $H_L \times W_L$ before being processed by the network.
Architectures with only local correlations (Local-Net) or with a unique global correlation (Global-Net) are visualized in Figure~\ref{arch}a, b.

\subsection{Global-Local Architecture}
\label{subsec:Mixed Global-Local}

We introduce a unified network that leverages the advantages of both global and local correlation layers and which also circumvents the limitations of both.
Our goal is to handle any kind of geometric transformations - including large displacements - while achieving high precision for detailed and small displacements. This is performed by carefully integrating global and local correlation layers in a feature pyramid based network architecture. 

Inspired by DGC-Net~\cite{Melekhov2019}, we employ a global correlation layer at the coarsest level. The purpose of this layer is to handle the \emph{long-range correspondences}. Since these are best captured in the coarsest scale, only a single global correlation is needed.  
In subsequent layers, the dense flow field is refined by computing image feature similarity using local correlations. This allows \emph{precise} estimation of the displacements. 
Combining global and local correlation layers allows us to achieve robust and accurate prediction of both long and small-range motions. Such an architecture is visualized with GLOCAL-Net in Figure~\ref{arch}c. However, this network is still restricted to a certain input resolution. Next, we introduce a design strategy that circumvents this issue.

\subsection{Adaptive resolution}
\label{subsec:adaptive_reso}

As previously discussed, the global correlation layer imposes a pre-determined input resolution for the network to ensure a constant channel dimensionality of the global cost volume. This severely limits the applicability and accuracy of the correspondence network, since higher resolution images requires down-scaling before being processed by the network, followed by up-scaling of the resulting flow. In this section, we address this key issue by introducing an architecture capable of taking images of \emph{any} resolution, while still benefiting from a global correlation.

Our adaptive-resolution architecture consists of two sub-networks, which operate on two different image resolutions. The first, termed L-Net, takes source and target images downscaled to a fixed resolution $H_{L} \times W_{L}$, which allows a global correlation layer to be integrated. The H-Net on the other hand, operates directly on the original image resolution $H \times W$, which is not constrained to any specific value. It refines the flow estimate generated by the L-Net with local correlations applied to a shallow feature pyramid constructed directly from the original images. It is schematically represented in Figure~\ref{arch}d.

Both sub-networks are based on a coarse-to-fine architecture, employing the same feature extractor backbone. In detail, the L-Net relies on a global correlation at the coarsest level in order to effectively handle any kind of geometric transformations, including very large displacements. Subsequent levels of L-Net employ local correlations to refine the flow field. It is then up-sampled to the coarsest resolution of H-Net, where it serves as the initial flow estimate used for warping the source features $F_\text{s}$. Subsequently, the flow prediction is refined numerous times within H-Net, that operates on the full scale images, thus providing a very detailed, sub-pixel accurate final estimation of the dense flow field relating $I_\text{s}$ and $I_\text{t}$. 

For high-resolution images, the upscaling factor between the finest pyramid level, $l_{L}$, of L-Net and the coarsest, $l_{H}$, of H-Net (see Figure~\ref{arch}d) can be significant. Our adaptive resolution strategy allows additional refinement steps of the flow estimate between those two levels during inference, thus improving the accuracy of the estimated flow, without training any additional weights. This is performed by recursively applying the $l_H$ layer weights at intermediate resolutions obtained by down-sampling the source and target features from $l_H$. 
In summary, our adaptive resolution network is capable of seamlessly predicting an accurate flow field in the original input resolution, while also benefiting from robustness to long-range correspondences provided by the global layer. The entire network is trained end-to-end.

\begin{figure*}[t]
\centering
\includegraphics[width=0.99\textwidth]{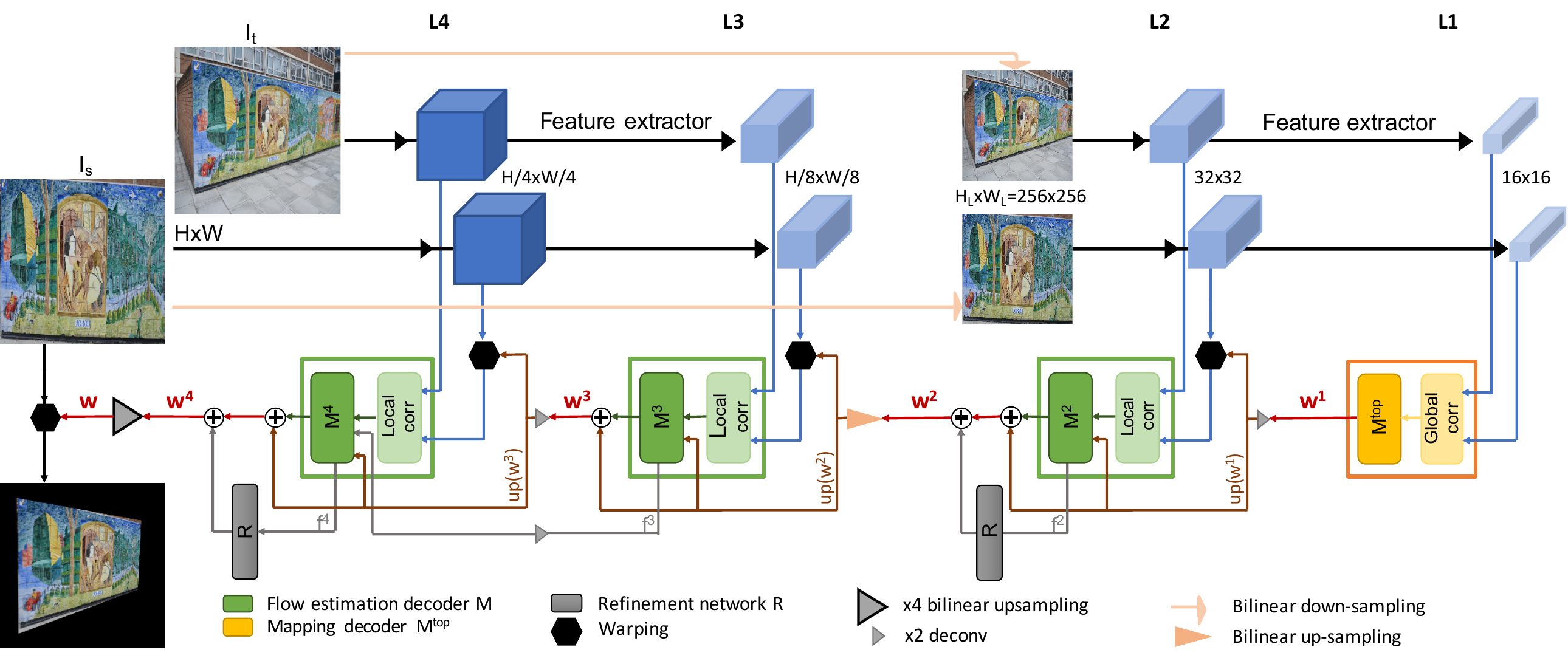}
\caption{Architectural details of our GLU-Net. It is composed of two modules, operating on two different image resolutions. The L-Net (right) relies on a global correlation for long-range matches, while the H-Net (left) refines the flow estimate with local correlations.}\vspace{-4mm}
\label{nets}
\end{figure*}

\subsection{Architecture details}
\label{subsec:architecture-det}

In this section, we provide a detailed description of our architecture. While any feature extractor backbone can be employed, we use the VGG-16~\cite{Chatfield14} network trained on ImageNet~\cite{Hinton2012} to provide a fair comparison to previous works in geometric~\cite{Melekhov2019} and semantic correspondences~\cite{Jeon}.
For our L-Net, we set the input resolution to $(H_{L} \times W_{L})=(256 \times 256)$. It is composed of two pyramid levels, using \verb|Conv5-3| ($16 \times 16$ resolution) and \verb|Conv4-3| ($32 \times 32$ resolution) respectively. The former employs global correlation, while the latter is based on a local correlation. The H-Net is composed of two feature pyramid levels extracted from the original image resolution $H \times W$. For this purpose, we employ \verb|Conv4-3|  and \verb|Conv3-3| having resolutions $\frac{H}{8} \times \frac{W}{8}$ and $\frac{H}{4} \times \frac{W}{4}$ respectively. The H-Net is purely based on local correlation layers. Our final architecture GLU-Net, composed of four pyramid levels in total, is detailed in Figure~\ref{nets}. Next, we describe the various architectural components.

\parsection{Coarsest resolution and mapping estimation} We compute a global correlation from the $L^2$-normalized source and target features. The cost volume is further post-processed by applying channel-wise $L^2$-normalisation followed by ReLU~\cite{relu} to strongly down-weight ambiguous matches~\cite{Rocco2017a}. Similar to DGC-Net~\cite{Melekhov2019}, the resulting global correlation C is then fed into a correspondence map decoder $M_\text{top}$ to estimate a 2D dense correspondence map $\mathbf{m}$ at the coarsest level $L1$ of the feature pyramid:
\begin{equation}
\label{eq:mapping-decoder}
    \mathbf{m}^{1}=M_\text{top}\left(C\left( 
    F_\text{t}^{1}, F_\text{s}^{1}
    \right)\right) \,.
    \end{equation}
The correspondence map is then converted to a displacement field, as $\mathbf{w}^1(\mathbf{x}) = \mathbf{m}^1(\mathbf{x})-\mathbf{x}$. 

\parsection{Subsequent flow estimations} The flow is refined by local correlation modules. At level $l$, the flow decoder $M$ infers the residual flow $\Delta \mathbf{\tilde{w}}^{l}$ as,
\begin{equation}
\label{eq:flow-decoder}
\Delta \mathbf{\tilde{w}}^{l}=M\left(c\left(F_\text{t}^l, \widetilde{F}_\text{s}^l ; R\right), \operatorname{up}\!\left(\mathbf{w}^{l-1}\right) \right) \,.
\end{equation}
$c$ is a local correlation \eqref{eq:local-corr} with search radius $R$ and $\widetilde{F}_\text{s}^l(\mathbf{x}) = F_\text{s}^l\left(\mathbf{x}+\operatorname{up}\left(\mathbf{w}^{l-1}\right )(\mathbf{x})\right)$ is the warped source feature map $F_\text{s}$ according to the upsampled flow $\operatorname{up}\!\left(\mathbf{w}^{l-1}\right )$. 
The complete flow field is computed as $\mathbf{\tilde{w}}^{l}=\Delta \mathbf{\tilde{w}}^{l} + \operatorname{up}\!\left(\mathbf{w}^{l-1}\right )$.

\parsection{Flow refinement} Contextual information have been shown advantageous for pixel-wise prediction tasks~\cite{Chen2017, DCCNet}. We thus use a sub-network R, called the refinement network, to post-process the estimated flow at the highest levels of L-Net and H-Net (L2 and L4 in Figure~\ref{nets}) by effectively enlarging the receptive field size. It takes the features $f^{l}$ of the second last layer from the flow decoder $M^{l}$ as input and outputs the refined flow $\mathbf{w}^{l}=R\left(f^{l}\right) + \mathbf{\tilde{w}}^{l}$. For the other pyramid level (L3), the final flow field is   $\mathbf{w}^{l}=\mathbf{\tilde{w}}^{l}$.

\parsection{Cyclic consistency} Since the quality of the correlation is of primary importance for the flow estimation process, we introduce an additional filtering step on the global cost volume to enforce the reciprocity constraint on matches. We employ the soft mutual nearest neighbor filtering introduced by~\cite{Rocco2018b} and apply it to post-process the global correlation. 

\subsection{Training}
\label{subsec:training}

\parsection{Loss}
We train our network in a single phase. We freeze the pre-trained backbone feature extractor during training. Following FlowNet~\cite{Dosovitskiy2015}, we apply supervision at every pyramid level using the endpoint error (EPE) loss with respect to the ground truth displacements.

\parsection{Dataset} Our network is solely trained on pairs generated by applying random warps to the original images.
Since our network is designed to also estimate correspondences between high-resolution images, training data of sufficient resolution is preferred in order to utilize the full potential of our architecture. 
We use a combination of the
DPED~\cite{Ignatov2017}, CityScapes~\cite{Cordts2016} and ADE-20K~\cite{Zhou2019} datasets, which have images larger than $750 \times 750$.
On the total dataset of $40,000$ images, we apply the same synthetic transformations as in \textit{tokyo} (DGC-Net~\cite{Melekhov2019} training data).  The resulting image pairs are cropped to $520 \times 520$ for training. We call this dataset \textit{DPED-CityScape-ADE}. 
More training and architectural details are available in the appendix Sections~\ref{sec:arc details} and~\ref{sec:training}.

\begin{figure*}[t!]
\centering
\vspace{-4mm}
\includegraphics*[width=0.99\textwidth]{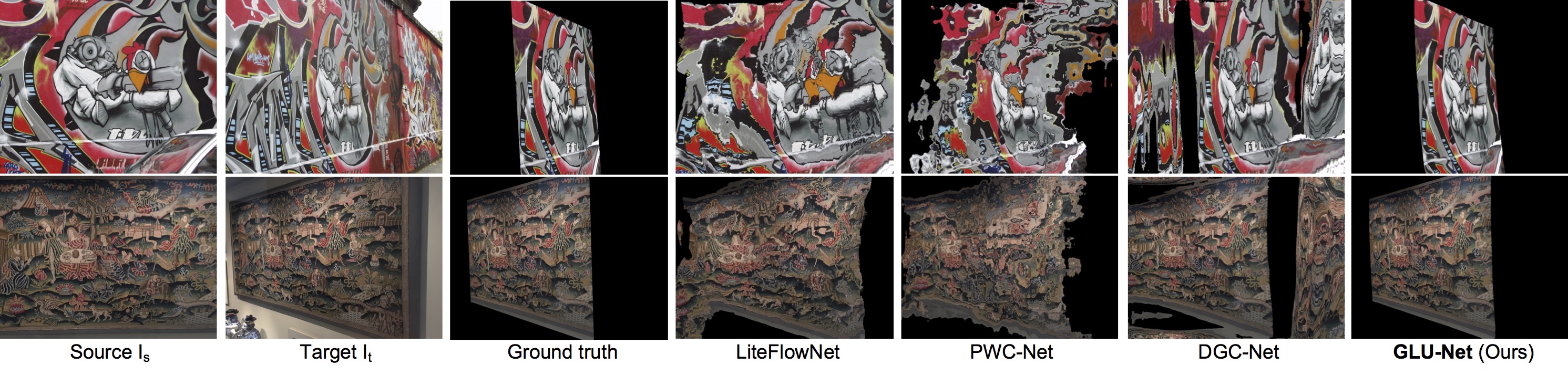} \\
\includegraphics*[width=0.99\textwidth]{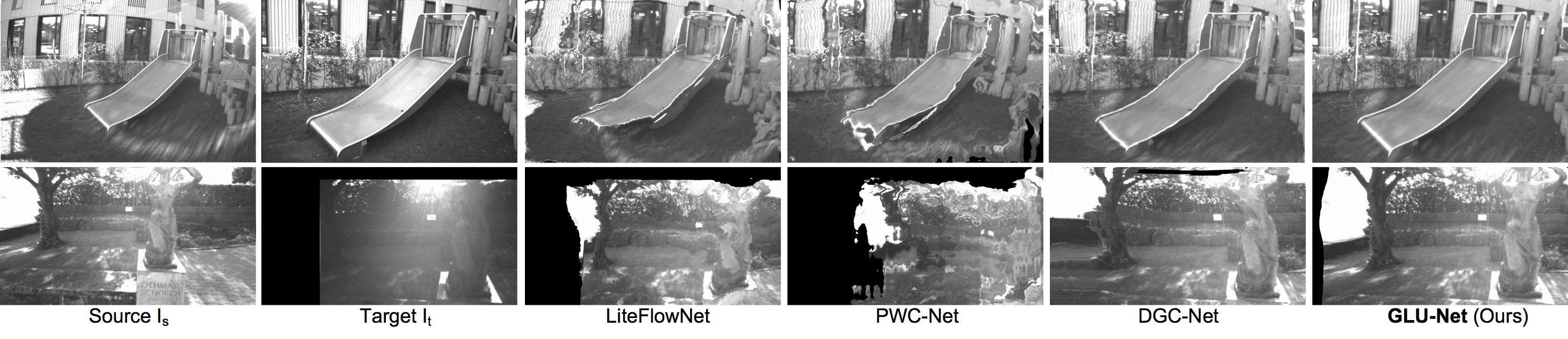} 
\vspace{-2mm}\caption{Qualitative comparison with state-of-the-art on geometric correspondence datasets. Top: Pairs of HP images. Bottom: Pairs of images from ETH3D taken by two different cameras. Our approach effectively handles large variations in view-point and appearance.}\vspace{-4mm}
\label{hp-images}
\end{figure*}
\section{Experimental Validation}
\label{sec:exp-val}

In this section, we comprehensively evaluate our approach for three diverse problems: geometric matching, semantic correspondences and optical flow. Importantly, we use the \emph{same} network and model weights, trained on \emph{DPED-CityScape-ADE}, for all three applications. More detailed results are available in the appendix Section~\ref{sec:results}. 

\subsection{Geometric matching}
\label{subsec:geo-match}

We first apply our GLU-Net for the task of geometric matching. 
The images thus consist of different views of the same scene and include large geometric transformations.

\parsection{HP} As in DGC-Net~\cite{Melekhov2019}, we employ the 59 sequences of the HPatches dataset~\cite{Lenc} labelled with \verb|v_X|, which have viewpoint changes, thus excluding the ones labelled \verb|i_X|, which only have illumination changes. Each image sequence contains a source image and 5 target images taken under increasingly larger viewpoints changes, with sizes ranging from $450 \times 600$ to $1613 \times 1210$. In addition to evaluating on the original image resolution (referred to as \textbf{HP}), we also evaluate on downscaled ($240 \times 240$) images and ground-truths (\textbf{HP-240}) following~\cite{Melekhov2019}. 

\parsection{ETH3D} To validate our approach for real 3D scenes, where image transformations are not constrained to simple homographies, we also employ the Multi-view dataset ETH3D~\cite{ETH3d}. It contains 10 image sequences at $480 \times 752$ or $514 \times 955$ resolution, depicting indoor and outdoor scenes.
The authors additionally provide a set of sparse geometrically consistent image correspondences (generated by~\cite{SchonbergerF16}) that have been optimized over the entire image sequence using the reprojection error. 
We sample image pairs from each sequence at different intervals to analyze varying magnitude of geometric transformations, and use the provided points as sparse ground truth correspondences. This results in about 500 image pairs in total for each selected interval. 

\parsection{Metrics} In line with \cite{Melekhov2019}, we employ the Average End-Point Error (AEPE) and Percentage of Correct Keypoints (PCK) as the evaluation metrics. AEPE is defined as the Euclidean distance between estimated and ground truth flow fields, averaged over all valid pixels of the target image. PCK is computed as the percentage of correspondences $\mathbf{\tilde{x}}_{j}$ with an Euclidean distance error $\left \| \mathbf{\tilde{x}}_{j} - \mathbf{x}_{j}\right \| \leq  \delta$, w.r.t.\ to the ground truth $\mathbf{x}_{j}$, that is smaller than a threshold $\delta$.

\parsection{Compared methods} We compare with DGC-Net~\cite{Melekhov2019} trained on \textit{tokyo}, which is the current state-of-the-art for dense geometric matching.
For a fair comparison we also train a version, called DGC-Net$^\dagger$, using the same data (\textit{DPED-CityScape-ADE}) as our GLU-Net.
We additionally compare with two state-of-the-art optical flow methods, PWC-Net~\cite{Sun2018} and LiteFlowNet~\cite{Hui2018}, both trained on \textit{Flying-Chairs}~\cite{Dosovitskiy2015} followed by \textit{3D-things}~\cite{Ilg2017a}. We use the PyTorch~\cite{pytorch} implementations~\cite{Melekhov2019, pytorch-liteflownet, Sun2018} of the models and the pre-trained weights provided by the authors.

\begin{table}[b]
\centering
\vspace{-5mm}\resizebox{0.49\textwidth}{!}{%
\begin{tabular}{lrrrr|rrr}
\toprule
                 & \multicolumn{4}{c}{\textbf{HP-240x240}}    & \multicolumn{3}{c}{\textbf{HP}}   \\ 
                 & Run-time & AEPE  & PCK-1px & PCK-5px & AEPE  & PCK-1px & PCK-5px    \\\midrule
LiteFlowNet~\cite{Hui2018}      &  45.10 ms & 19.41     &    28.36 \%        &    57.66  \%      &  118.85    &   13.91     \%  &  31.64  \%    \\
PWC-Net~\cite{Sun2018, Sun2019}  & 38.51 ms & 21.68 & 20.99 \% & 54.19 \% & 96.14 & 13.14 \%  & 37.14  \% \\
DGC-Net~\cite{Melekhov2019}    & 138.30 ms & 9.07  & 50.01 \%      & 77.40 \%      & 33.26 & 12.00 \%   & 58.06 \% \\
DGC-Net$^\dagger$ & 138.30 ms & 9.12 & 43.09 \% & 79.35 \% & 33.47 & 9.19 \% & 56.02 \% \\
\textbf{GLU-Net} (Ours)    &  \textbf{38.10} ms & \textbf{7.40} &    \textbf{59.92} \%   &    \textbf{83.47} \%   &  \textbf{25.05} &  \textbf{39.55} \%  &  \textbf{78.54} \% \\ \bottomrule
\end{tabular}%
}\vspace{1mm}
\caption{Comparison of state-of-the-art algorithms applied to the task of geometric matching, on the HPatches dataset~\cite{Lenc}. Lower AEPE and higher PCK are better.}\vspace{-2mm}
\label{tab:geo-match-HP}
\end{table}

\parsection{Results} We first present results on the HP and HP-240 in Table~\ref{tab:geo-match-HP}. Our model strongly outperforms all others by a large margin both in terms of accuracy (PCK) and robustness (AEPE). 
It is interesting to note that while our model is already better than DGC-Net on the small resolution HP-240, the gap in performance further broadens when increasing the image resolution. Particularly, GLU-Net obtains a PCK-1px value almost four times higher than that of DGC-Net on HP. This demonstrates the benefit of our adaptive resolution strategy, which enables to process high-resolution images with high accuracy. Moreover, our model achieves a 3.6 times faster inference compared to DGC-Net. 
Figure~\ref{hp-images} shows qualitative examples of different networks applied to HP images and ETH3D image pairs taken by two different cameras. Our GLU-Net is robust to large view-points variations as well as drastic changes in illumination.

\begin{figure}[t]
\vspace{-6mm}\includegraphics[width=0.47\textwidth,trim=3 15 0 0]{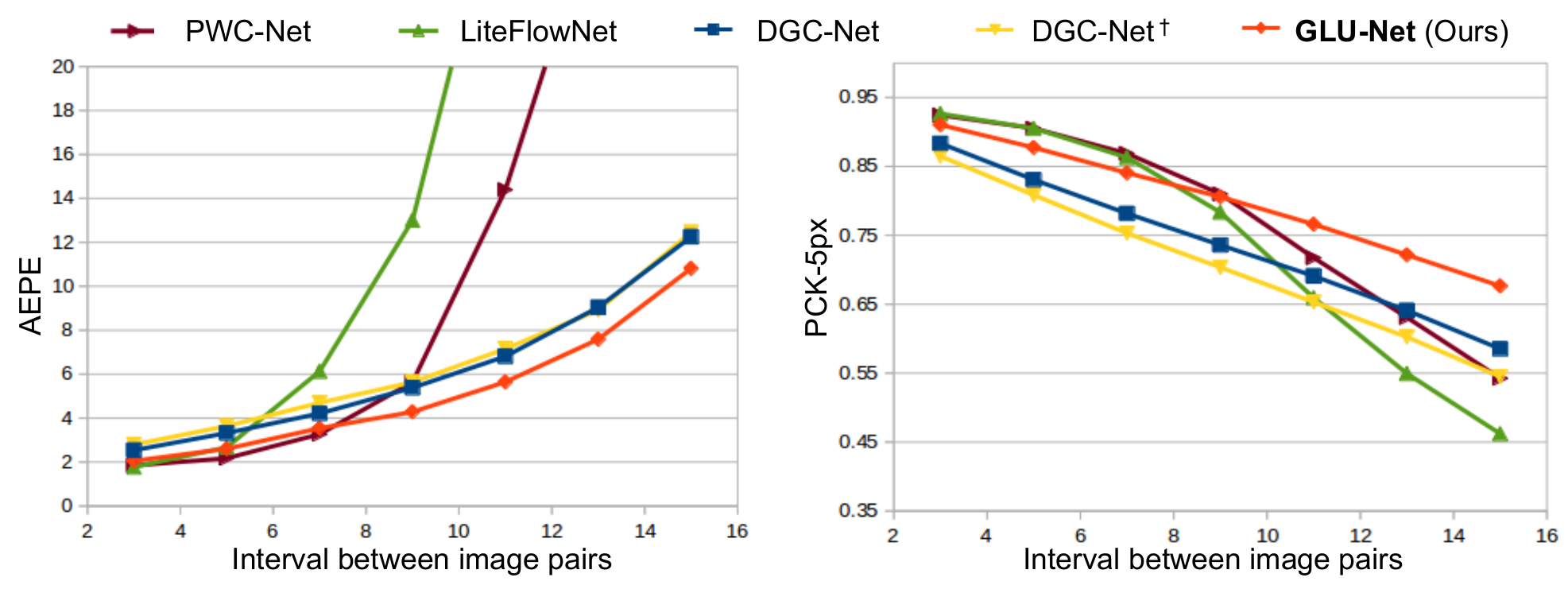}
\caption{Quantitative results on geometric matching dataset ETH3D~\cite{ETH3d}. AEPE and PCK-5 are computed on pairs of images sampled from consecutive images of ETH3D at different intervals.}\vspace{-5mm}
\label{ETH3d}
\end{figure}

In Figure~\ref{ETH3d}, we plot AEPE and PCK-5px obtained on the ETH3D scenes for different intervals between image pairs. For small intervals, finding correspondences strongly resembles optical flow task while increasing it leads to larger displacements. Therefore, specialised optical flow methods PWC-Net~\cite{Sun2018} and LiteFlowNet~\cite{Hui2018} obtain slightly better AEPE and PCK for low intervals, but rapidly degrade for larger ones. 
In all cases, our approach consistently outperforms DGC-Net~\cite{Melekhov2019} in both metrics by a large margin.

\subsection{Semantic matching}
\label{subsec:sem-match}

Here, we perform experiments for the task of semantic matching, where images depict different instances of the same object category, such as cars or horses. We use the same model and weights as in the previous section.

\parsection{Dataset and metric} We use the TSS dataset~\cite{Taniai2016}, which provides dense flow fields annotations for the foreground object in each pair. It contains 400 image pairs, divided into three groups: FG3DCAR, JODS, and PASCAL, according to the origins of the images. Following Taniai~\etal~\cite{Taniai2016}, we report the PCK with a distance threshold equal to $\alpha \cdot \max(H_{s}, W_{s})$, where $H_{s}$ and $W_{s}$ are the dimensions of the source image and $\alpha = 0.05$.

\parsection{Compared methods} We compare to several recent state-of-the-art methods specialised in semantic matching \cite{DCCNet,Jeon, Kim2018,Kim2019,Rocco2018a, Rocco2018b}. 
In addition to our universal network, we evaluate a version that adopts two architectural details that are used in the semantic correspondence literature. Specifically, we add a consensus network~\cite{Rocco2018b} for the global correlation layer and concatenate features from different levels in the L-Net, similarly to~\cite{Jeon} (see Section~\ref{sec:ablation} for an analysis). 
We call this version \emph{Semantic-GLU-Net}. To accommodate reflections, which do not occur in geometric correspondence scenarios, we infer the flow field on original and flipped versions of the target image and output the flow field with least horizontal average magnitude.

\begin{figure}[b]
\centering
\vspace{-4mm}
\includegraphics[width=0.48\textwidth,trim=0 18 0 0]{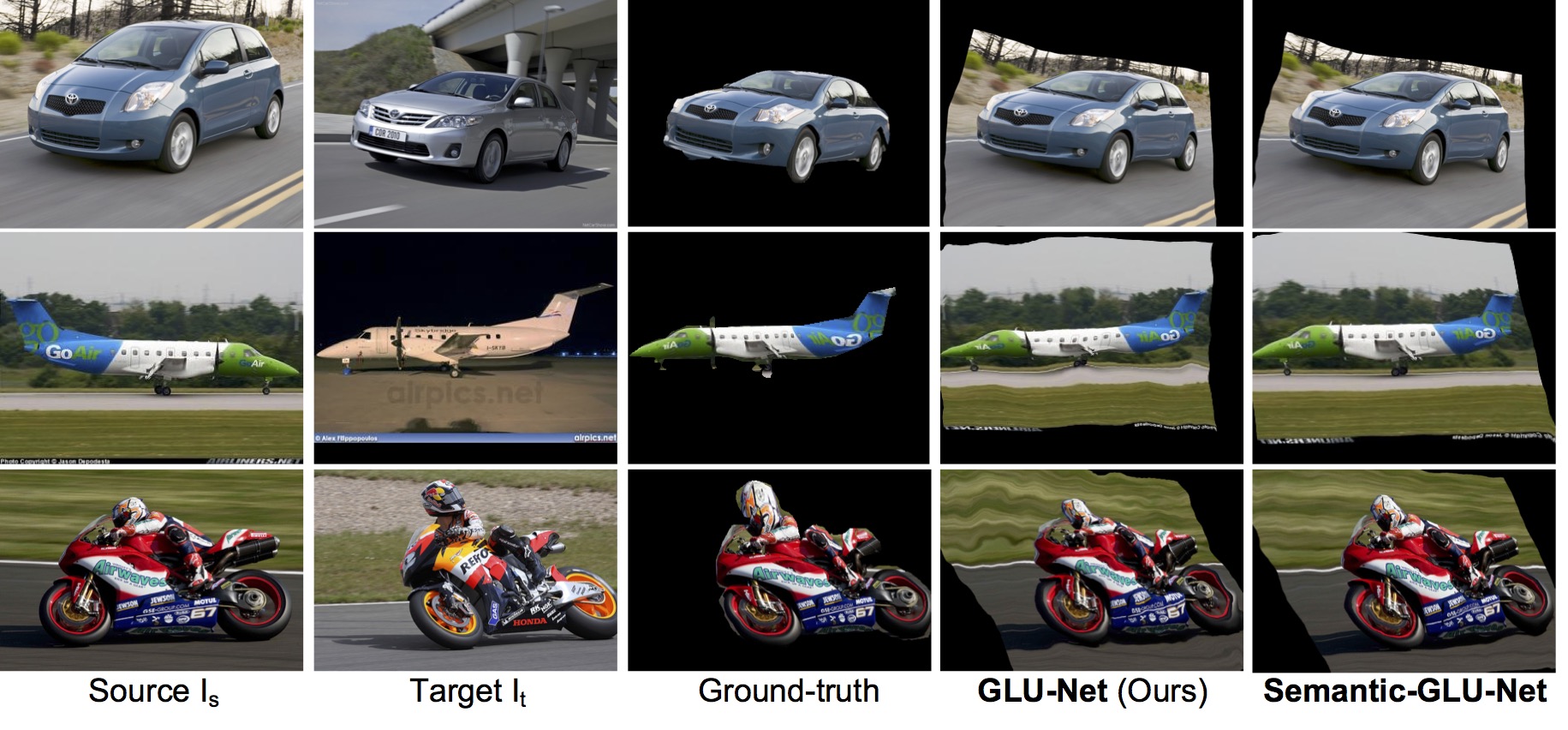}
\caption{Qualitative examples of GLU-Net (Ours) and Semantic-GLU-Net applied to TSS images~\cite{Taniai2016}.}\vspace{-2mm}
\label{TSS-fig}
\end{figure}

\parsection{Results} We report results on TSS in Table~\ref{tab:TSS}. Our universal network obtains state-of-the-art performance on average over the three TSS groups. Moreover, individual results on FG3Dcar and PASCAL are very close to best metrics. This shows the generalization properties of our network, which is not trained on the same magnitude of semantic data. 
In contrast, most specialized approaches fine-tuned on PASCAL data~\cite{Ham2016}. Finally, including architectural details specifically for semantic matching, termed Semantic-GLU-Net, further improves our performance, setting a new state-of-the-art on TSS, by improving a substantial $1.0\%$ PCK over the previous best. Interestingly, we outperform methods that use a deeper, more powerful feature backbone. 
Qualitative examples of our approach are shown in Figure~\ref{TSS-fig}.

\begin{table}[t]
\centering
\vspace{-4mm}\resizebox{0.48\textwidth}{!}{%
\begin{tabular}{llcccc}
\toprule
Methods  & Feature backbone &  FG3DCar & JODS & PASCAL & Avg. \\ \midrule
CNNGeo(W)  
\cite{Rocco2018a}& ResNet-101 & 90.3  & 76.4 & 56.5  & 74.4 \\
RTNs \cite{Kim2018} & ResNet-101 & 90.1 & 78.2 & 63.3 & 77.2 \\
PARN \cite{Jeon} & VGG-16 & 87.6 & 71.6 & 68.8 & 76.0 \\
PARN  \cite{Jeon} & ResNet-101 & 89.5  & 75.9 & 71.2  & 78.8 \\
NC-Net 
\cite{Rocco2018b} &  ResNet-101 & 94.5 & 81.4 & 57.1 & 77.7 \\
DCCNet \cite{DCCNet}  & ResNet-101 & 93.5  & \textbf{82.6} & 57.6  & 77.9 \\
SAM-Net \cite{Kim2019} & VGG-19  & \textbf{96.1} & 82.2 & 67.2 & 81.8 \\ \midrule
\textbf{GLU-Net} & VGG-16   &   93.2    &   73.3  &   71.1    &    79.2  \\ 
\textbf{Semantic-GLU-Net} & VGG-16 & 94.4 & 75.5  & \textbf{78.3} & \textbf{82.8} \\ \bottomrule
\end{tabular}%
}\vspace{1mm}
\caption{PCK [\%] obtained by different state-of-the-art methods on TSS~\cite{Taniai2016} for the task of semantic matching. 
}\vspace{-5mm}
\label{tab:TSS}
\end{table}

\subsection{Optical flow}
\label{subsec:OF}

Finally, we apply our network, with the same weights as previously, for the task of optical flow estimation. Here, the image pairs stem from consecutive frames of a video.

\parsection{Dataset and metric} For optical flow evaluation, we use the KITTI dataset~\cite{Geiger2013}, which is composed of real road sequences captured by a car-mounted stereo camera rig. The 2012 set only consists of static scenes while the 2015 set is extended to dynamic scenes.  For this task, we follow the standard evaluation metric, namely the Average End-Point Error (AEPE). We also use the KITTI-specific F1 metric, which represents the percentage of outliers.

\parsection{Compared methods} We employ state-of-the-art PWC-Net~\cite{Sun2018, Sun2019} and LiteFlowNet~\cite{Hui2018} trained on \textit{Flying-Chairs}~\cite{Dosovitskiy2015} and \textit{3D-things}~\cite{Ilg2017a}. We also compare to DGC-Net~\cite{Melekhov2019} (\textit{tokyo}) and DGC-Net$^\dagger$ (\textit{DPED-CityScape-ADE}).

\begin{table}[t]
\centering
\vspace{-2mm}\resizebox{0.40\textwidth}{!}{%
\begin{tabular}{lcc|cc}
\toprule
              & \multicolumn{2}{c}{\textbf{KITTI-2012}} & \multicolumn{2}{c}{\textbf{KITTI-2015}} \\ 
              & AEPE-all               & F1-all   [\%]         & AEPE-all                & F1-all  [\%]              \\ \midrule
LiteFlowNet~\cite{Hui2018}   &    4.00                 &          \textbf{17.47}$^*$         &       10.39                &        \textbf{28.50}          \\       
PWC-Net~\cite{Sun2018, Sun2019}     &    4.14                   &      20.28$^*$             &      10.35               &             33.67      \\
DGC-Net~\cite{Melekhov2019}      &       8.50$^*$              &   32.28$^*$                &       14.97$^*$              &        50.98$^*$           \\
DGC-Net$^\dagger$ & 7.96 &  34.35 & 14.33 & 50.35 \\
\textbf{GLU-Net} &        \textbf{3.34}             &      18.93             &      \textbf{9.79}               &        37.52           \\ \bottomrule
\end{tabular}%
}\vspace{1mm}
\caption{Quantitative results on optical flow KITTI training datasets~\cite{Geiger2013}. Fl-all: Percentage of outliers averaged over all pixels. Inliers are defined as $AEPE <3$ pixels or $<5 \%$. Lower F1 and AEPE are best. $^*$ Denotes values which are computed using the trained models provided by the authors.}\vspace{-5mm}
\label{tab:KITTI}
\end{table}

\begin{figure}[b]
\centering
\vspace{-4mm}
\includegraphics[width=0.48\textwidth,trim=0 10 0 0]{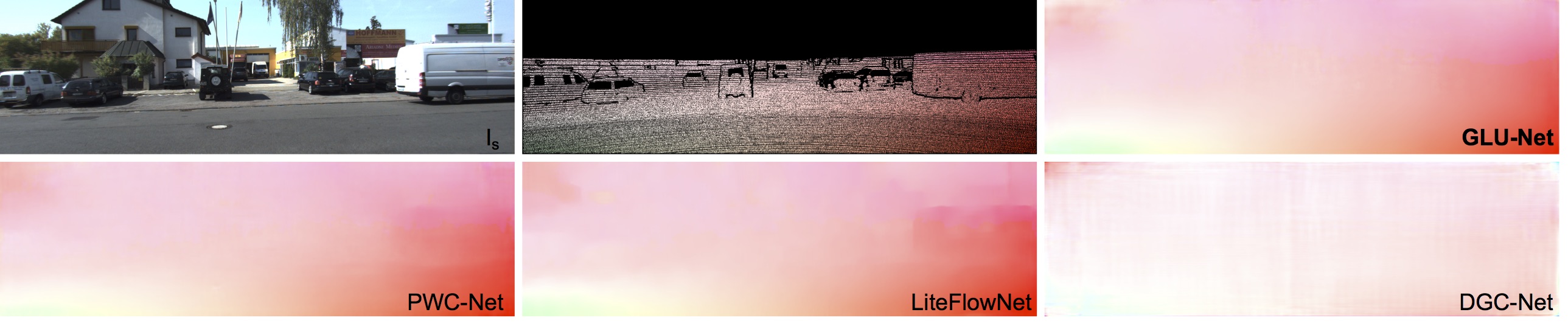}
\caption{Visualization of the flow outputted by different methods for a KITTI-2012 image.}\vspace{-2mm}
\label{kitti}
\end{figure}

\parsection{Results} Since we do not finetune our model, we only evaluate on the KITTI training sets. For fair comparison, we compare to models not finetuned on the KITTI training data. The results are shown in Table~\ref{tab:KITTI} and a qualitative example is illustrated in Figure~\ref{kitti}. Our network obtains highest AEPE on both KITTI-2012 and 2015. Nevertheless, we observe that our method achieves a larger F1 on KITTI-2015 compared to approaches specifically trained and designed for optical flow. This is largely due to our self-supervised training data, which currently does not model independently moving objects or occlusions, but could be included to pursue a more purposed optical flow solution. Yet, our approach demonstrates competitive results for this challenging task, without training on any optical flow data. This clearly shows that our network can not only robustly estimate long-range matches, but also accurate small displacements.

\subsection{Ablation study}
\label{sec:ablation}

Here, we perform a detailed analysis of our approach.

\parsection{Local-global architecture} We first analyze the impact of global and local correlation layers in our dense correspondence framework. We compare using only local layers (Local-Net), a global layer (Global-Net) and our combination (GLOCAL-Net), presented in Figure~\ref{arch}. As shown in Table~\ref{tab:ablation-a}, Local-Net fails on the HP dataset, due to its inability to capture large displacements. While the Global-Net can handle large viewpoint changes, it achieves inferior accuracy compared to GLOCAL-Net, which additionaly integrates local correlations layers.

\parsection{Adaptive resolution} By further adding the adaptive resolution strategy (Section~\ref{subsec:adaptive_reso}), our approach (GLU-Net in Table~\ref{tab:ablation-a}) achieves a large performance gain in all metrics compared to GLOCAL-Net. This improvement is most prominent for high resolution images, \ie the original HP data.

\parsection{Iterative refinement} From Table~\ref{tab:ablation-a}, applying iterative refinement (\textbf{it-R}) clearly benefits accuracy for high-resolution images (HP). 
This further allows us to seamlessly add extra flow refinements, without incurring any additional network weights, in order to process images of high resolution.

\parsection{Global correlation} Lastly, we explore design choices for the global correlation block in our architecture. As shown in Table~\ref{tab:ablation-b}, adding cyclic consistency (\textbf{CC})~\cite{Rocco2018b} as a post-processing brings improvements for all datasets.
Subsequently adding \textbf{NC-Net} and concatenating features of L-Net (\textbf{Concat-F}) lead to major overall gain on the HP~\cite{Lenc} and TSS~\cite{Taniai2016} datasets. However, we observe a slight degradation in accuracy, as seen on KITTI~\cite{Geiger2013}. We therefore only include these components for the Semantic-GLU-Net version (Section~\ref{subsec:sem-match}) and not in our universal GLU-Net.

\begin{table}[t]
\centering
\vspace{-2mm}\resizebox{0.48\textwidth}{!}{%
\begin{tabular}{llccc|cc}
\toprule
 && Local- & Global-   & GLOCAL- & \textbf{GLU-Net} & \textbf{GLU-Net}\\
 && Net &           Net   &      Net  & (no CC, no it-R) & (no CC, it-R) \\ \midrule
 
\multirow{3}{6mm}{\textbf{HP-240}}
&AEPE             &   10.62 & 9.72 & \textbf{8.77} & 7.69 & \textbf{7.69}\\ 
&PCK-1px [\%] &   35.10 & 41.28 &  \textbf{48.53} & 53.83 & \textbf{53.83}\\ 
&PCK-5px [\%] &  73.03 & 72.76 &  \textbf{78.12} & 83.17 & \textbf{83.17}\\ \midrule
\multirow{3}{6mm}{\textbf{HP}}
&AEPE             &   147.96 & 34.64 &\textbf{31.64} & 25.55 & \textbf{25.09}\\ 
&PCK-1px [\%] &  7.41  &  8.86 &\textbf{10.23} & 35.26 & \textbf{36.81}\\ 
&PCK-5px [\%] & 19.27 &  50.11 &  \textbf{56.73} & 75.79 & \textbf{77.55} \\ \bottomrule
\end{tabular}%
}\vspace{1mm}
\caption{Effect of global and local correlations as well of adaptive resolution strategy. \textbf{it-R}: iterative refinement, introduced with our adaptive resolution (Section~\ref{subsec:adaptive_reso}), \textbf{CC}: cyclic-consistency~\cite{Rocco2018b}.}\vspace{-1.5mm}
\label{tab:ablation-a}
\end{table}

\begin{table}[t]
\centering
\resizebox{0.48\textwidth}{!}{%
\begin{tabular}{llcccc}
\toprule
 &&  No CC & + CC (Ours) & + NC-Net & + Concat-F \\ \midrule
\multirow{3}{*}{\textbf{HP}}
&AEPE & 25.09  &  25.05  & 22.00 &  \textbf{21.40}\\ 
&PCK-1px [\%] & 36.81 &  \textbf{39.55}   & 37.62 & 38.49 \\ 
&PCK-5px [\%] & 77.55 &  78.54    & 79.41 & \textbf{79.50}\\ \midrule
\textbf{KITTI-} & AEPE & 3.56  &  \textbf{3.34}    & 3.80 & 3.85\\ 
\textbf{2012} & F1-all {[}\%{]} & 21.67  &   \textbf{18.93}     & 23.49 & 23.84\\ \midrule
\textbf{TSS} 
&PCK [\%] & 78.97  & 79.21 & 82.10 & \textbf{82.76}\\ \bottomrule
\end{tabular}%
}\vspace{1mm}
\caption{Effect of additional architectural details. All models are with iterative refinement. We add  \textbf{CC}: cyclic-consistency~\cite{Rocco2018b}, \textbf{NC-Net}: Neighborhood Consensus network~\cite{Rocco2018b}, \textbf{Concat-F}: Concatenation of features of L-Net~\cite{Jeon}.}\vspace{-5mm}
\label{tab:ablation-b}
\end{table}
\section{Conclusion}
\label{sec:conclusion}

We propose a universal coarse-to-fine architecture for estimating  dense flow fields from a pair of images. By carefully combining global and local correlation layers, our network effectively estimates long-range displacements while also achieving high accuracy. Crucially, we introduce an adaptive resolution strategy to counter the fixed input resolution otherwise imposed by the global correlation. Our universal GLU-Net is thoroughly evaluated for the three diverse tasks of geometric correspondences, semantic matching and optical flow. When using the same model weights, our network achieves state-of-the-art performance on all above tasks, demonstrating its universal applicability.

\noindent\textbf{Acknowledgments:}
This work was partly supported by the ETH Z\"urich Fund (OK), a Huawei Technologies Oy (Finland) project, an Amazon AWS grant, and Nvidia.

{\small
\bibliographystyle{ieee_fullname}
\bibliography{biblio}
}

\newpage
\clearpage
\appendix

\begin{center}
	\textbf{\Large Appendix}
\end{center}


In this supplementary material, we first provide details about the architecture of the different modules of our network GLU-Net in Section~\ref{sec:arc details}. We then explain the training procedure in more depth in Section~\ref{sec:training}. Finally, we present additional qualitative results and more detailed quantitative experiments in Section~\ref{sec:results}.

\section{Architecture details}
\label{sec:arc details}

In this section, we provide additional details about cyclic consistency as a post processing step of the global correlation. We also give a detailed architectural description of the mapping and flow decoders, along  with the refinement network. Lastly, we explain in depth the iterative refinement allowed by our adaptive resolution strategy. In the following, a convolution layer or block refers to the composition of a 2D-convolution followed by batch norm~\cite{IoffeS15} and ReLU~\cite{relu} (Conv-BN-ReLU).

\subsection{Cyclic consistency post-processing step for improved global correlation} 
Since the quality of the correlation layer output is of primary importance for the flow estimation process, we introduce an additional filtering step on the global cost volume to enforce the reciprocity constraint on matches. To encourage matched features to be mutual nearest neighbours, we employ the soft mutual nearest neighbor filtering introduced by~\cite{Rocco2018b} and apply it to post-process the global correlation. 

The soft mutual nearest neighbor module filters a global correlation $C \in \mathbb{R}^{H \times W \times H \times W}$ into $\hat{C} \in \mathbb{R}^{H \times W \times H \times W}$ such that:
\begin{equation}
\hat{C}(i, j, k, l)=r_t(i, j, k, l) \cdot r_s(i, j, k, l) \cdot C(i, j, k, l)
\end{equation}
with $ r_s(i, j, k, l)$ and $ r_t(i, j, k, l)$ the ratios of the score of the particular match $C(i, j, k, l)$ with the best scores along each pair of dimensions corresponding to images $I_\text{s}$ and $I_\text{t}$ respectively. We present the formula for $r_s(i, j, k, l)$ below, the same applies for $ r_t(i, j, k, l)$.

\begin{equation}
r_t(i, j, k, l)=\frac{C(i, j, k, l)}{\max _{a b} C(a, b, k, l)}
\end{equation}
This cyclic consistency post-processing step does not add any training weights. 

\subsection{Mapping decoder $M_\text{top}$}

In this sub-section, we give additional details of the mapping decoder $M_\text{top}$ for the global correlation layer (Eq.~\ref{eq:mapping-decoder} and Figure~\ref{nets} in the paper).
We compute a global correlation from the $L^2$-normalized source and target features. The cost volume is further post-processed by applying channel-wise $L^2$-normalisation followed by ReLU~\cite{relu} to strongly down-weight ambiguous matches~\cite{Rocco2017a}. Similar to DGC-Net~\cite{Melekhov2019}, the resulting global correlation layer C is then fed into a correspondence map decoder $M_\text{top}$ to estimate a 2D dense correspondence map $\mathbf{m}$ at the coarsest level $L_1$ of the feature pyramid,
\begin{equation}
\mathbf{m}^{1}=M_\text{top}\left(C\left( 
\frac{F_\text{t}^{1}}{\left \| F_\text{t}^{1} \right \|}, \frac{F_\text{s}^{1}}{\left \| F_\text{s}^{1} \right \|}
\right)\right) \,.
\end{equation}
The outputted mapping estimate is parameterized such that each predicted pixel location in the map belongs to the interval $\left [-1;1  \right ]$  representing width and height normalized image coordinates. 
The correspondence map is then re-scaled to image coordinates and converted to a displacement field. 
\begin{equation}
\mathbf{w}^1(\mathbf{x}) = \mathbf{m}^1(\mathbf{x}) - \mathbf{x} \,.
\end{equation}
The decoder $M_{top}$ consists of 5 feed-forward convolutional blocks with a $3 \times 3$ spatial kernel. The number of feature channels of each convolutional layers are respectively 128, 128, 96, 64, and 32. The final output of the mapping decoder is the result of a linear 2D convolution, without any activation.

\subsection{Flow decoder $M$} 
Here, we give additional details of the flow decoder $M$ for the local correlation layers (Eq.~\ref{eq:flow-decoder} and Figure~\ref{nets} in the paper). At level $l$, the flow decoder $M$ infers the residual flow $\Delta \mathbf{\tilde{w}}^{l}$ as,
\begin{equation}
\Delta \mathbf{\tilde{w}}^{l}=M\left(c\left(F_\text{t}^l, \widetilde{F}_\text{s}^l ; R\right), \operatorname{up}(\mathbf{w}^{l-1}) \right) \,.
\end{equation}
Here, $c$ is a local correlation volume with search radius $R$ and $\widetilde{F}_\text{s}^l(\mathbf{x}) = F_\text{s}^l\left(\mathbf{x}+\operatorname{up}\left(\mathbf{w}^{l-1}\right )(\mathbf{x})\right)$ is the warped source feature map $F_\text{s}$ according to the upsampled flow from the previous pyramid level $\operatorname{up}\left(\mathbf{w}^{l-1}\right )$. The complete flow field is then computed as $\mathbf{\tilde{w}}^{l}=\Delta \mathbf{\tilde{w}}^{l} + \operatorname{up}\left(\mathbf{w}^{l-1}\right )$.

The flow decoder at level 4 (see Figure~\ref{nets} of main paper) additionally takes an input $\operatorname{de}_2(f^{l-1})$, obtained by applying a transposed convolution layer $\operatorname{de}_2$ to the features $f^{l-1}$ of the second last layer from the flow decoder $M^{l-1}$.  This additional inputs was first introduced and utilized in PWC-Net~\cite{Sun2018} at every pyramid level. It enables the decoder of the current level to obtain some information about the correlation at the previous level. In GLU-Net, this additional input to the flow decoder only appears in H-Net since in L-Net, a global correlation and mapping decoder precede the flow decoder. 

As for the flow decoder $M$, we employ a similar architecture to the one in PWC-Net~\cite{Sun2018}. It consists of 5 convolutional layers with DenseNet connections~\cite{Huang2017}. The numbers of feature channels at each convolutional layers are respectively 128, 128, 96, 64, and 32, and the spatial kernel of each convolution is $3 \times 3$. DenseNet connections are used since they have been shown to lead to significant improvement in image classification~\cite{Huang2017} and optical flow estimation~\cite{Sun2018}. The final output of the flow decoder is the result of a linear 2D convolution, without any activation.

\subsection{Refinement network $R$} 
Here, we explain in more details the refinement network R (Figure~\ref{nets} in the paper). The refinement network aims to refine the pixel-level flow field $\mathbf{\tilde{w}}^{l}$, thus preventing erroneous flows from being amplified by up-sampling and passing to the next pyramid level. Its architecture is the same than the context network employed in PWC-Net~\cite{Sun2018}. It is a feed-forward CNN with 7 dilated convolutional layers~\cite{YuK15}, with varying dilation rates. Dilated convolutions enlarge the receptive field without increasing the number of weights. From bottom to top, the dilation constants are 1, 2, 4, 8, 16, 1, and 1. The spatial kernel is set to $3 \times 3$ for all convolutional layers.

\subsection{Details about Local-net, Global-Net and GLOCAL-Net}
In Figure~\ref{arch} of the main paper, we introduced Local-Net, Global-Net and GLOCAL-Net to investigate the differences between architectures based on local correlation layers, a global correlation layer or a combination of the two, respectively. All three networks are composed of three pyramid levels and use the same feature extractor backbone VGG-16~\cite{Chatfield14}. The mapping and flow decoders have the same architecture as those used for GLU-Net and described above. For Global-Net, the pyramid levels following the global correlation level employ concatenation of the target and warped source feature maps, as suggested in DGC-Net~\cite{Melekhov2019}. They are fed to the flow estimation decoder along with the up-sampled flow from the previous resolution. Finally, Global-Net and GLOCAL-Net are both restricted to a pre-determined input resolution $H_L \times W_L$ due to their global correlation at the coarsest pyramid level. On the other hand, Local-Net, which only relies on local correlations, can take input images of any resolutions. 

\subsection{Iterative refinement}

\begin{figure*}
\centering
\includegraphics[width=0.70\textwidth]{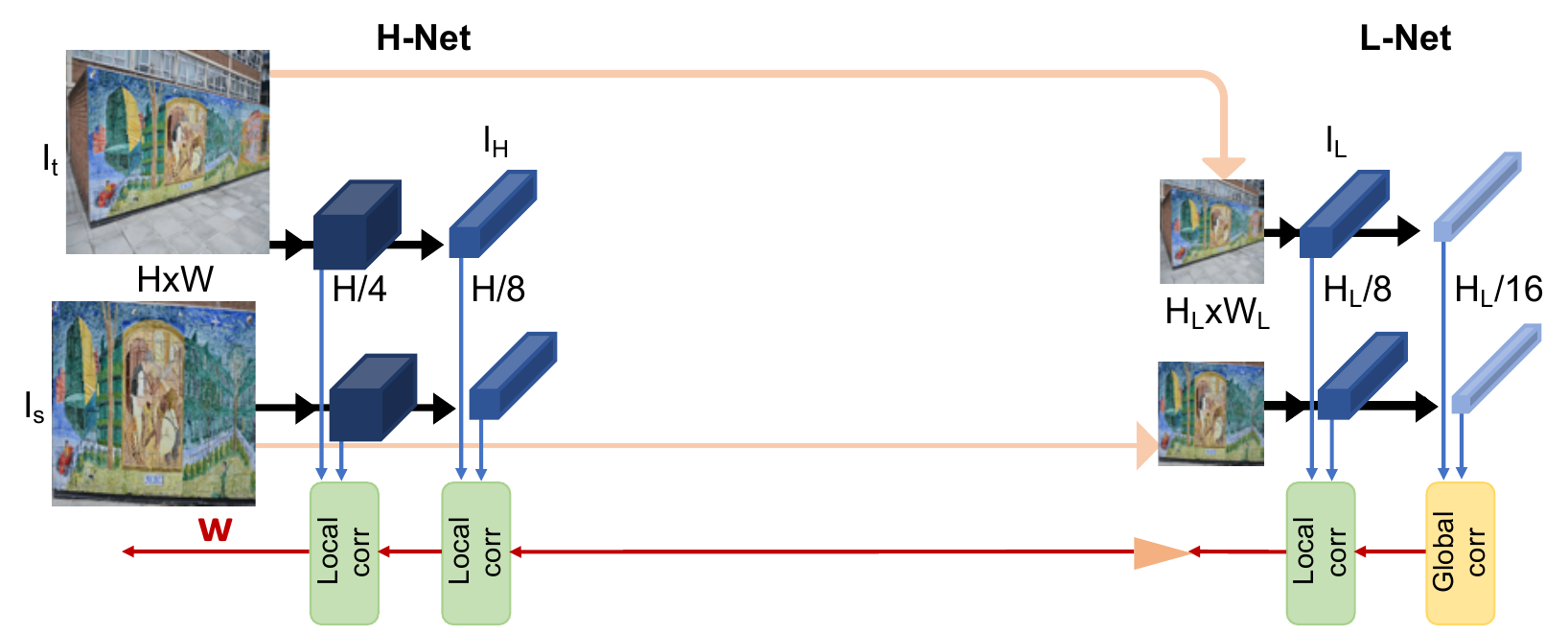} \\
(a) GLU-Net without iterative refinement. \\
\vspace{3mm}\includegraphics[width=0.70\textwidth]{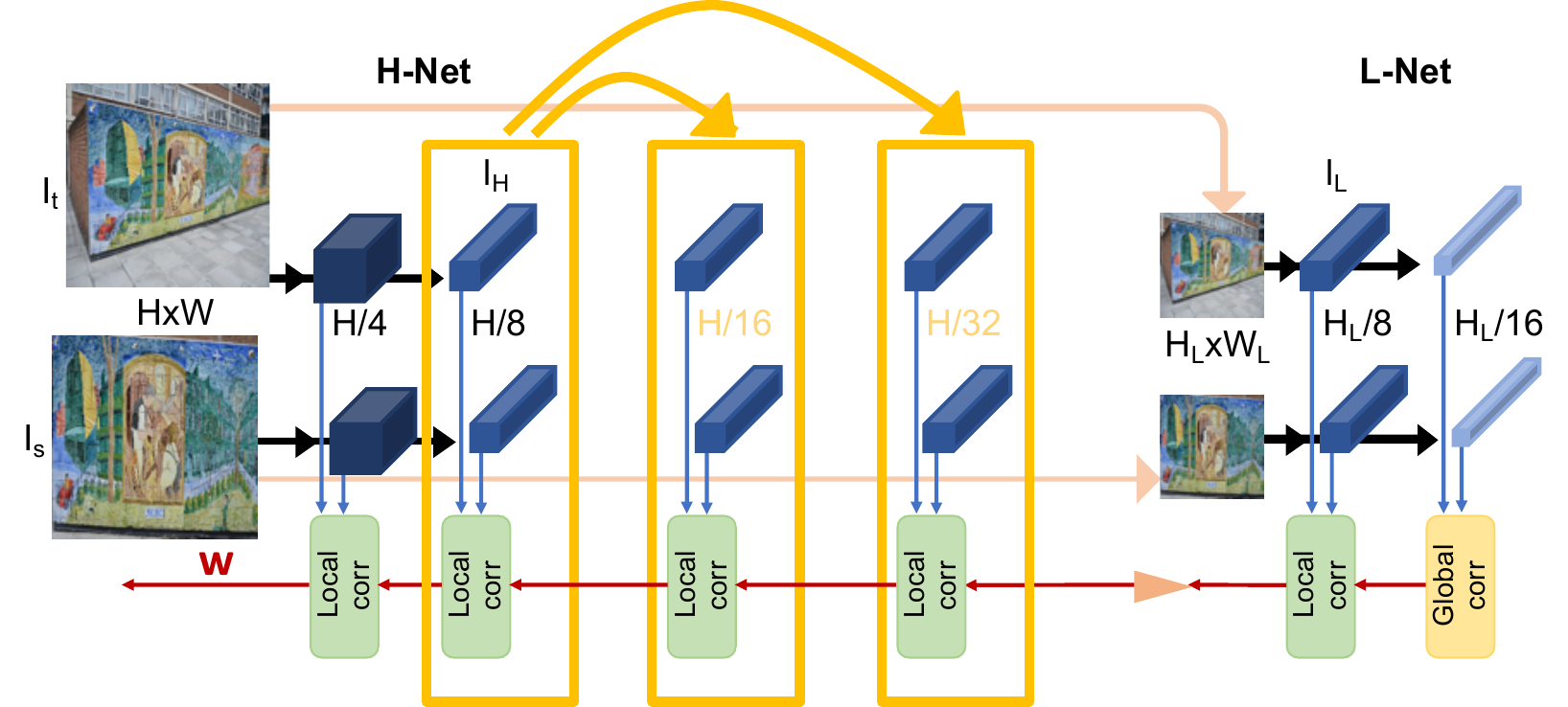} \\
(b) GLU-Net with iterative refinement between L-Net and H-Net. \\
\vspace{3mm}\caption{Schematic representation of iterative refinement. The features and weights of $l_H$ level of H-Net are iteratively applied at intermediate resolutions between L-Net and H-Net.}
\label{it-R}
\end{figure*}

Here we provide more details about the iterative refinement procedure described in~\ref{subsec:adaptive_reso} in the paper. For high-resolution images, the upscaling factor between the finest pyramid level, $l_{L}$, of L-Net and the coarsest, $l_{H}$, of H-Net (see Figure~\ref{it-R}) can be significant. Our adaptive resolution strategy allows additional refinement steps of the flow estimate between those two levels during inference, thus improving the accuracy of the estimated flow, without training any additional weights. This is performed by recursively applying the $l_H$ layer weights at intermediate resolutions obtained by down-sampling the source and target feature maps from $l_H$. 

Particularly, we apply iterative refinement if the ratio between the resolutions of the $l_H$ and $l_L$ levels is larger than three. We then iteratively perform refinements at intermediate resolutions, obtained by a reduction of factor 2 from $l_H$ in each step, until the ratio between the resolution of the coarsest intermediate level and the resolution of $l_L$ is smaller than 2.

In more details, we construct a local correlation layer from the source and target feature maps of level $l_H$ down-sampled to the desired intermediate resolution. We then apply the weights of the level $l_H$ decoder to the local correlation, therefore obtaining an intermediate refinement of the flow field. 
This process is illustrated in Figure~\ref{it-R}, where the gap between $l_L$ and $l_H$ here allows for two additional flow field refinements.

\section{Training details}
\label{sec:training}

Here, we provide additional details about the training procedure and the training dataset. 

\subsection{Loss} 
We freeze the weights of the feature extractor during training. 
Let $\theta$ denote the learnable parameters of the network. 
Let $\mathbf{w}_{\theta}^l=(\mathbf{w_x}^l, \mathbf{w_y}^l) \in \mathbb{R}^{H_{l} \times W_l \times 2}$ denote the flow field estimated by the network at the $l^{th}$ pyramid level. $\mathbf{w}_\mathrm{GT}^l$ refers to the corresponding dense flow ground-truth, computed from the random warp. We employ the multi-scale training loss, first introduced in FlowNet~\cite{Dosovitskiy2015}, 
\begin{equation}
    \mathcal{L}(\theta)=\sum_{l=L_{1}}^{L} \alpha_{l} 
    \sum_{\mathbf{x}} \left\| \mathbf{w}_{\theta}^{l}(\mathbf{x})-\mathbf{w}_{\mathrm{GT}}^{l}(\mathbf{x}) \right\| +\gamma \left\| \theta \right\|\,,
\end{equation}
where  $\alpha_{l}$ are the weights applied to each pyramid level and the second term of the loss regularizes the weights of the network. 
We do not apply any mask during training, which means that occluded regions (that do not have visible matches) are included in the training loss. Since the image pairs are related by synthetic transformations, these regions do have a correct ground-truth flow value. 

For our adaptive resolution strategy, we down-sample and scale the ground truth from original resolution $H\times W$ to $H_L \times W_L$ in order to obtain the ground truth flow fields for L-Net. Similarly to FlowNet~\cite{Dosovitskiy2015} and PWC-Net~\cite{Sun2018}, we down-sample the ground truth from the base resolution to the different pyramid resolutions without further scaling, so as to obtain the supervision signals at the different levels. 

\subsection{Dataset} To use the full potential of our GLU-Net, training should be performed on high-resolution images. We create the training dataset following the procedure in DGC-Net~\cite{Melekhov2019}, but enforcing the condition of high resolution. We use the same $40,000$ synthetic transformations (affine, thin-plate and homographies), but apply them to our higher resolution images collected from the DPED~\cite{Ignatov2017}, CityScapes~\cite{Cordts2016} and ADE-20K~\cite{Zhou2019} datasets. Indeed, DPED images are very large, however the DPED training dataset is composed of only approximately 5000 sets of images taken by four different cameras. We use the images from two cameras, resulting in around  $10,000$ images. CityScapes additionally adds about $23,000$ images. We complement with a random sample of ADE-20K images with a minimum resolution of $750 \times 750$.

\subsection{Implementation details} As a preprocessing step, the training images are mean-centered and normalized using mean and standard deviation of ImageNet dataset~\cite{Hinton2012}. For all local correlation layers, we employ a search radius $R=4$. For the training of Global-Net and GLOCAL-Net, which both have a pre-determined fixed input image resolution of $(H_L \times W_L = 256 \times 256)$, we use a batch size of 32 while we train LOCAL-Net, which can take any input image, with batches of size 16. We set the initial learning rate to $10^{-2}$ and gradually decrease it during training. The weights in the training loss are set to be $\alpha_{1}=0.32, \alpha_{2}=0.08, \alpha_{3}=0.02$.

Our final network GLU-Net is trained with a batch size of 16 and the learning rate initially equal to $10^{-4}$. The weights in the training loss are set to be $\alpha_{1}=0.32, \alpha_{2}=0.08, \alpha_{3}=0.02, \alpha_{4}=0.01$.
Our system is implemented using Pytorch~\cite{pytorch} and our networks are trained using Adam optimizer~\cite{adam} with learning rate decay of $0.0004$.

\begin{table*}[t]
\centering
\vspace{-3mm}\resizebox{0.99\textwidth}{!}{%
\begin{tabular}{lccc|ccc|cc|cc}
\toprule
                 & \multicolumn{3}{c}{\textbf{HP-240x240}}    & \multicolumn{3}{c}{\textbf{HP}}   &
                 \multicolumn{2}{c}{\textbf{KITTI-2012}} & \multicolumn{2}{c}{\textbf{KITTI-2015}} \\
                & AEPE  & PCK-1px [\%] & PCK-5px [\%] & AEPE  & PCK-1px [\%] & PCK-5px [\%]  &  AEPE-all    & F1-all  [\%] &  AEPE-all      & F1-all  [\%] \\ \midrule
                
DGC-Net (\textit{tokyo}) & 9.07  & 50.01      & 77.40       & 33.26 & 12.00   & 58.06  & 8.50 & 32.38 & 14.97 & 50.98 \\

DGC-Net$^\dagger$ (\textit{DPED-CityScape-ADE)}  & 9.12 & 43.09 & 79.35  & 33.47 & 9.19 & 56.02  & 7.96 &  34.35 & 14.33 & 50.35 \\

\textbf{GLU-Net} (\textit{DPED-CityScape-ADE}) & \textbf{7.40} &    \textbf{59.92}    &    \textbf{83.47}   &  \textbf{25.05} &  \textbf{39.55}  &  \textbf{78.54} &  \textbf{3.34} & \textbf{18.93} &  \textbf{9.79}               &        \textbf{37.52}           \\ \bottomrule
\end{tabular}%
}\vspace{1mm}
\caption{Effect of the training dataset on the evaluation results of DGC-Net and comparison to GLU-Net. The training dataset is indicated in parenthesis.}\vspace{-1mm}
\label{tab:training-dat}
\end{table*}

\begin{table*}[t]
\centering
\resizebox{\textwidth}{!}{%
\begin{tabular}{@{}ll|llllll|llllll@{}}
\toprule
 & & \multicolumn{5}{c}{\textbf{HP-240}} & \multicolumn{5}{c}{\textbf{HP}} \\ \midrule
 & & I  & II  & III  & IV  & V  & all & I  & II & III & IV & V & all\\
                      &AEPE  & 6.99   &  16.78   & 19.13     & 25.27    & 28.89   & 19.41   &   36.69 & 102.17 & 113.58 & 154.97 & 186.82 & 118.85 \\
 \textbf{LiteFlowNet} & PCK-1px [\%] &  50.06  &    28.93 & 25.87     & 23.22    &  13.72  &    28.36 &  34.86 & 12.95 & 10.35 & 6.93 & 4.47 & 13.91 \\
                      & PCK-5px [\%] &  82.14  &     59.62 & 56.92      &  51.04   &    38.59 & 57.66   &   63.99 & 32.88 & 28.99 & 18.52 & 13.85 & 31.64 \\ \midrule

  & AEPE & 5.74 & 17.69 & 20.46 & 27.61 & 36.97 & 21.68 & 23.93 & 76.33 & 91.30  & 124.22 & 164.91 & 96.14 \\
\textbf{PWC-Net}  & PCK-1px [\%] & 43.55 & 20.35 & 18.60 & 14.17 & 8.27 &   20.99 & 31.56 & 12.10 & 10.83 & 7.09 & 4.12 & 13.14 \\
& PCK-5px [\%] & 80.06 & 57.08 & 53.89 & 45.70 & 34.22 & 54.19 & 68.79 & 38.51 & 36.38 & 25.24 & 16.76 & 37.14 \\ \midrule

                   & AEPE  & 1.74   & 5.88    & 9.07     &    12.14 & 16.50   & 9.07   & 5.71   & 20.48    & 34.15   & 43.94 & 62.01 & 33.26  \\
  \textbf{DGC-Net} & PCK-1px [\%] &   70.29 &  53.97   & 52.06     & 41.02    &    32.74 &  50.01    &     20.92 & 12.88   & 12.85 & 7.66 & 5.67 & 12.00  \\
                   &PCK-5px [\%] &   93.70 &  82.43   &  77.58    & 71.53    &    61.78& 77.40   &    78.88 & 63.37    & 60.21   & 48.83 & 38.99 & 58.06  \\ \midrule
                   
   & AEPE &  1.90 & 5.65 & 9.42 & 11.39 & 17.26 & 9.11 & 6.04 & 21.60 & 32.87 & 41.82 & 65.03 & 33.47 \\
   \textbf{DGC-Net$^\dagger$}  & PCK-1px [\%]  & 60.88 & 47.88 & 46.01 & 34.87 & 25.80 & 43.09 & 15.81 & 9.86 & 9.84 & 6.17 & 4.29 & 9.19 \\
   & PCK-5px [\%]  & 93.47 & 84.04 & 80.28 & 74.93 & 63.76 & 79.35 & 75.44 & 62.16 & 59.58 & 46.71 & 36.21 & 56.02 \\ \midrule
                   
               & AEPE  &  \textbf{0.59}  & \textbf{4.05}    & \textbf{7.64}     &  \textbf{9.82}   &    \textbf{14.89} & \textbf{7.40}   & \textbf{1.55}   & \textbf{12.66}    &    \textbf{27.54} &  \textbf{32.04} & \textbf{51.47} & \textbf{25.05}  \\
 \textbf{GLU-Net (Ours)} & PCK-1px [\%] & \textbf{87.89}   & \textbf{67.49}    &    \textbf{62.31}  &    \textbf{47.76} & \textbf{34.14}   & \textbf{59.92}   & \textbf{61.72}   &  \textbf{42.43}   &    \textbf{40.57} & \textbf{29.47} & \textbf{23.55} & \textbf{39.55}  \\
               &PCK-5px  [\%]& \textbf{99.14}  &    \textbf{92.39} & \textbf{85.87}     & \textbf{78.10}    & \textbf{61.84}   &    \textbf{83.47}& \textbf{96.15}   &    \textbf{84.35} & \textbf{79.46}   & \textbf{73.80} & \textbf{58.92} & \textbf{78.54}  \\  \bottomrule
\end{tabular}%
}\vspace{1mm}
\caption{Details of AEPE and PCK evaluated over each view-point ID of HP and HP-240 datasets.}
\label{tab:details-HP}
\end{table*}
\begin{figure*}[t]
\centering
\includegraphics[width=0.99\textwidth]{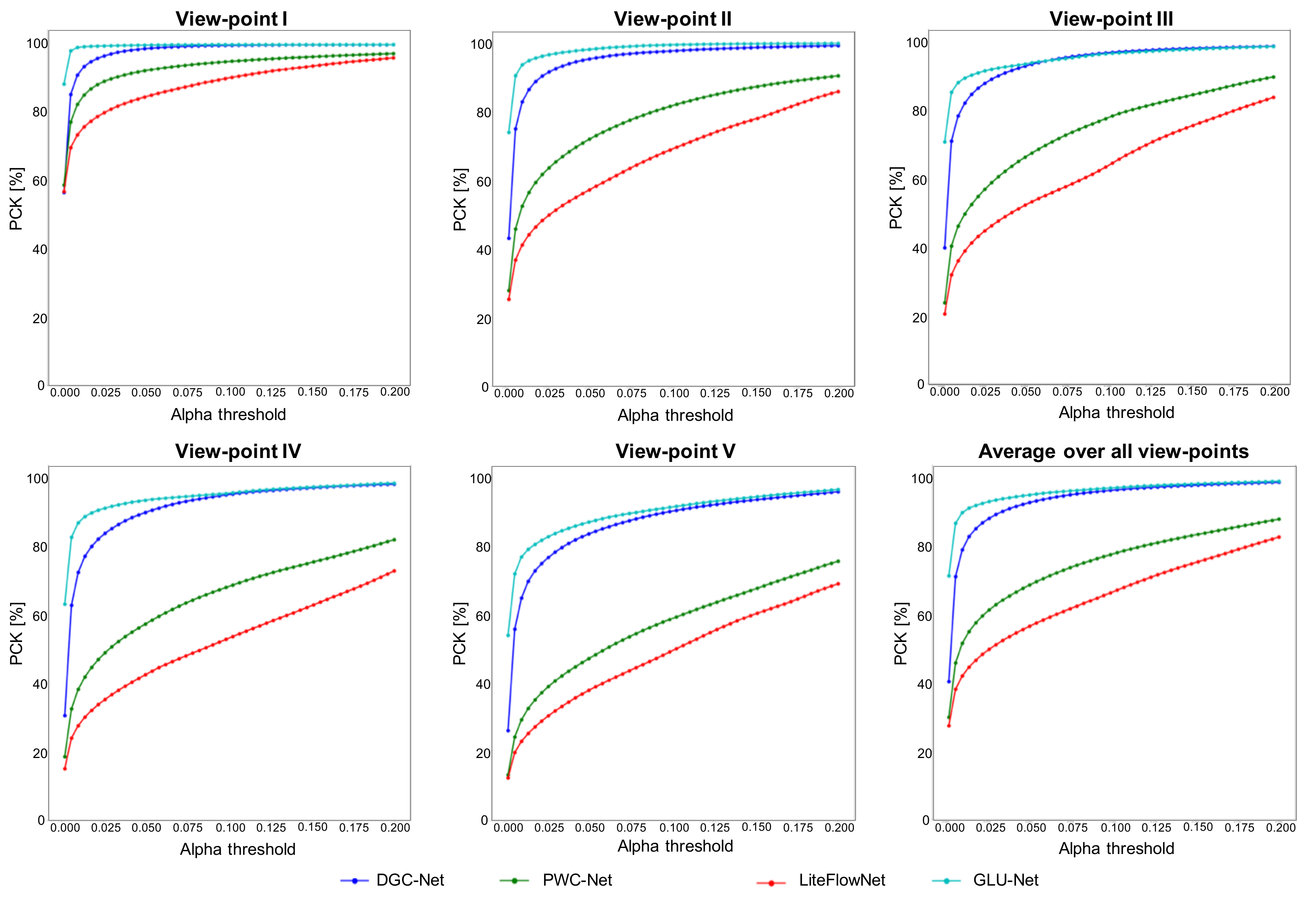}
\caption{PCK curves obtained by state-of-the-art methods and GLU-Net over the different view-points of HP. }
\label{pck}
\end{figure*}

\section{Detailed results}
\label{sec:results}

Here, we first provide additional details on the run-time computation in Section~\ref{run-time}. Then, in Section~\ref{training-dat}, we evaluate the influence of the training dataset on the evaluation results. We then present additional qualitative and more detailed quantitative results on  subsequently the geometric matching, the semantic matching and the optical flow tasks in respectively Sections~\ref{geo}, ~\ref{semantic} and~\ref{OF}. Finally, we expose additional ablation experiments in Section~\ref{ablation-study}.

\subsection{Run time}
\label{run-time}

We compare the run time of our method with state-of-the-art approaches over the HP-240 images in Table~\ref{tab:runtime}. The timings have been obtained on the same desktop with an NVIDIA GTX 1080 Ti GPU. The HP-240 images are of size $240 \times 240$, which corresponds to the pre-determined input resolution of DGC-Net. For PWC-Net, LiteFlowNet and GLU-Net, the images are resized to $256 \times 256$ before being passed through the networks. We do not consider this resizing in the estimated time. They all output a flow at a quarter resolution the input image. We up-scale to the image resolution $240 \times 240$ with bilinear interpolation. This up-scaling operation is included in the estimated time.

Our network GLU-Net obtains similar run time than PWC-Net and is three times faster than DGC-Net. This is due to the fact that PWC-Net, LiteFlowNet and GLU-Net outputs a flow at a quarter image resolution whereas DGC-Net refines the estimated flow field with two additional pyramid levels until the fixed resolution of $240 \times 240$. 

\begin{table}[b]
\centering
\resizebox{0.49\textwidth}{!}{%
\begin{tabular}{@{}lllll@{}}
\toprule
                & PWC-Net  & LiteFlowNet & DGC-Net & GLU-Net (Ours) \\\midrule
Run-time [ms]  & 38.51 & 45.10 & 138.30 & \textbf{38.10} \\ \bottomrule
\end{tabular}%
}\vspace{1mm}
\caption{Run time of our methods compared to optical-flow competitors PWC-Net and LiteFlowNet as well as geometric matching competitor DGC-Net, averaged over 295 image pairs of HP-240.}
\label{tab:runtime}
\end{table}

\subsection{Training dataset}
\label{training-dat}

Since DGC-Net is our main competitor, for a fair comparison, we additionally trained DGC-Net on our training dataset \textit{DPED-CityScape-ADE}, using the training code provided by the authors, which resulted in DGC-Net$^\dagger$. 
In Table~\ref{tab:training-dat}, we summarize the results of DGC-Net trained on both \textit{DPED-CityScape-ADE} or \textit{tokyo} and evaluated on geometric matching datasets HP-240 and HP as well as optical flow datasets KITTI-2012 and KITTI-2015. It seems that the training dataset in this case only has a small effect. Since both datasets were created by applying the same synthetic transformations, this support the fact that geometric transformation and displacement statistics are more important for generalization properties than image content~\cite{Mayer2018, Schuster2019b,Sun2019}.

\subsection{Geometric matching}
\label{geo}

We provide the detailed results on HP and ETH3D datasets, as well as extensive additional qualitative examples. We also analyse the performance of our network with respect to rotation and scaling.

\subsubsection{Results on HPatches dataset}

\begin{figure*}[t]
\centering
\includegraphics[width=0.99\textwidth]{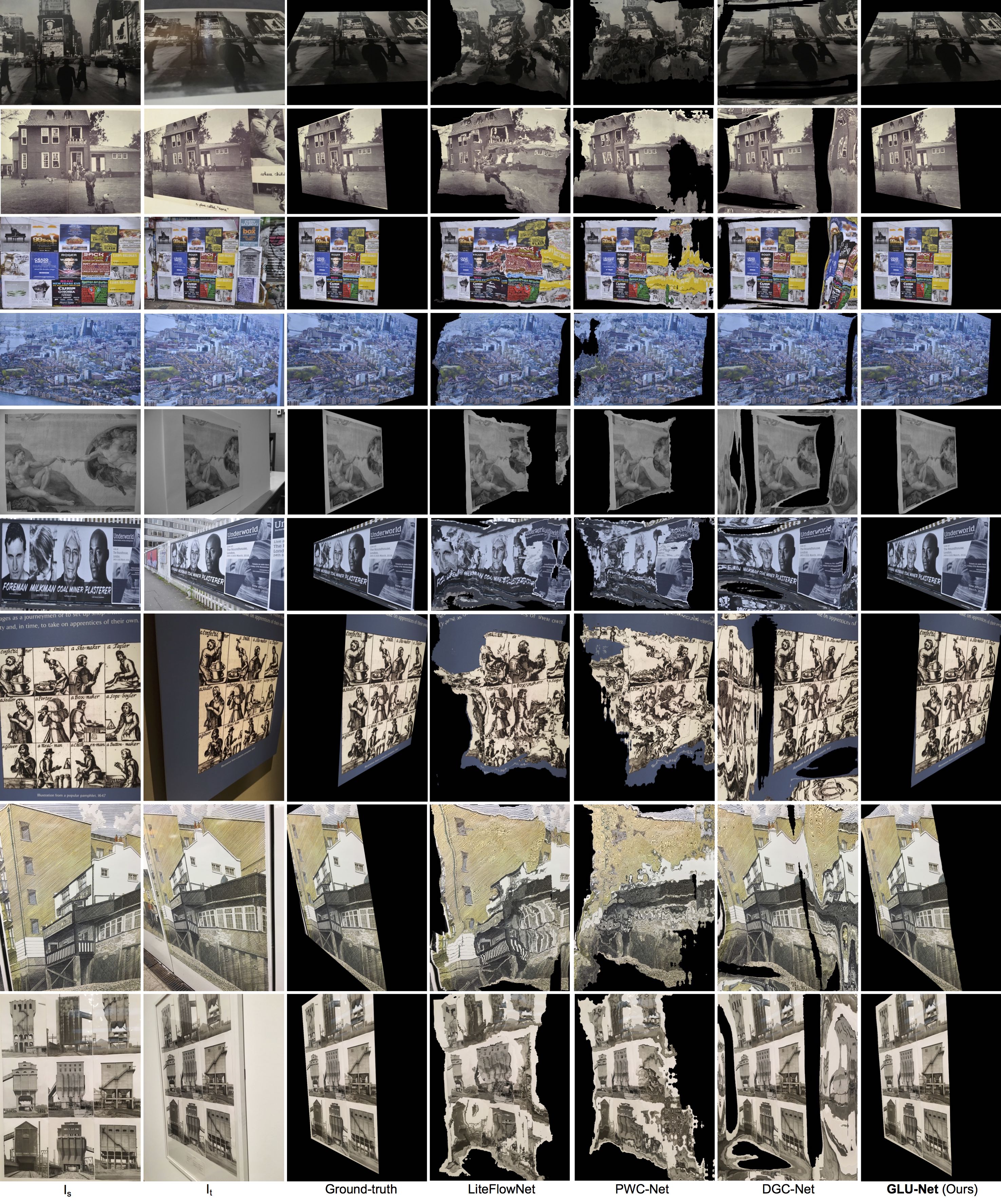}
\caption{Qualitative examples of different state-of-the-art algorithms and our GLU-Net applied to HP images. The source images are warped according to the flow fields outputted by the different networks. The warped source images should resemble the target images. Our method GLU-Net is robust to drastic view-point changes.}
\label{hp}
\end{figure*}

Detailed results obtained by different models on the various view-points of the HP and HP-240 datasets are presented in Table~\ref{tab:details-HP}. It corresponds to Table~\ref{tab:geo-match-HP} of the main paper, that only provides the average over all viewpoint IDs. Note that increasing view-point IDs lead to increasing geometric transformations due to larger changes in viewpoint.
We outperform all other methods for each viewpoint ID on both low resolution (HP-240) and high-resolution images (HP). Particularly, our network permits to gain a lot of accuracy (in the order of 3 to 4 times higher for PCK-1 on HP) as compared to DGC-Net. Additional qualitative examples are shown in Figure~\ref{hp}.

We additionally present the PCK curves computed over the different viewpoints of HP, as a function of the relative distance threshold. We do not set a pixel-level thresholds for the PCK curves since HP image pairs have different resolutions in general. GLU-Net achieves better accuracy (better PCK) for all thresholds compared to PWC-Net~\cite{Sun2018}, LiteFLowNet~\cite{Hui2018} and DGC-Net~\cite{Melekhov2019}. Importantly, GLU-Net obtains significantly better PCK for low thresholds.

\subsubsection{Results on ETH3D} 

In the main paper, Figure~\ref{ETH3d}, we quantitatively evaluated our approach over pairs of ETH3D images sampled from consecutive frames at different intervals. In Table~\ref{tab:ETH3d-details}, we give the corresponding detailed evaluation metrics (AEPE and PCK) obtained by PWC-Net, LiteFlowNet, DGC-Net, DGC-Net$^\dagger$ and GLU-Net. 

Note that the PCK and AEPE values computed per image are then averaged over all image pairs of each sequence. The final metrics presented are the averages over all sequences.

\begin{table}
\centering
\resizebox{0.49\textwidth}{!}{%
\begin{tabular}{llccccc}
\toprule
&&  LiteFlowNet & PWC-Net & DGC-Net & DGC-Net$^\dagger$ & GLU-Net (Ours) \\ \midrule
 
& AEPE &  \textbf{1.67} & 1.75 & 2.49 & 2.76 & 1.98 \\
\textbf{interval = 3} &PCK-1px [\%] & \textbf{61.63} & 57.54 & 34.19 & 27.55 & 50.55 \\
&PCK-5px [\%] & \textbf{92.79} & 92.62 & 88.50 & 86.65  & 91.22 \\ \midrule

&AEPE &  2.58 & \textbf{2.10} & 3.28 & 3.61 & 2.54 \\
\textbf{interval = 5} &PCK-1px [\%] & \textbf{56.55} & 50.41 & 27.22 & 21.23 & 43.08 \\
&PCK-5px [\%] & \textbf{90.70} & 90.71 & 83.25  & 81.01 & 87.91 \\ \midrule

&AEPE &  6.05 & \textbf{3.21} & 4.18 & 4.67 & 3.48 \\
\textbf{interval = 7} &PCK-1px [\%] & \textbf{49.83} & 42.95 & 22.45 & 16.91 & 36.98 \\
&PCK-5px [\%] & 86.29 & \textbf{87.04} & 78.32  & 75.50 & 84.23 \\ \midrule

&AEPE &         12.95 & 5.59 & 5.35 & 5.61 & \textbf{4.23} \\
\textbf{interval = 9} &PCK-1px [\%] & \textbf{42.00} & 35.23 & 18.82 & 13.91 & 32.45 \\
&PCK-5px [\%] & 78.50 & \textbf{81.17} &  73.74 & 70.54 &  80.74 \\ \midrule

&AEPE &         29.67 & 14.35 & 6.78 & 7.13 & \textbf{5.59} \\
\textbf{interval = 11} &PCK-1px [\%] & \textbf{33.14} & 28.14 & 15.82 & 11.67 & 28.45 \\
&PCK-5px [\%] & 66.07 & 71.91 & 69.23  & 65.48 & \textbf{76.84} \\ \midrule

&AEPE &         52.41 & 27.49 & 9.02 & 8.89 & \textbf{7.54} \\
\textbf{interval = 13} &PCK-1px [\%] & \textbf{26.46} & 22.91 & 13.49 & 9.82 & 25.06 \\
&PCK-5px [\%] & 55.05 & 63.19 & 64.28  & 60.42 & \textbf{72.35} \\ \midrule

&AEPE &         74.96 & 43.41 & 12.25 & 12.45 & \textbf{10.75} \\
\textbf{interval = 15} &PCK-1px [\%] & 21.22 & 18.34 & 11.25 & 8.14 & \textbf{21.89} \\
&PCK-5px [\%] & 46.29 & 54.39 & 58.66  & 54.69 & \textbf{67.77} \\ \midrule
\end{tabular}%
}
\caption{Metrics evaluated over scenes of ETH3D with different intervals between consecutive pairs of images (taken by the same camera). Note that those results are the average over the different sequences of the ETH3D dataset. Small AEPE and high PCK are better.}\label{tab:ETH3d-details}
\end{table}

\begin{figure*}[t]
\centering
\includegraphics[width=0.99\textwidth]{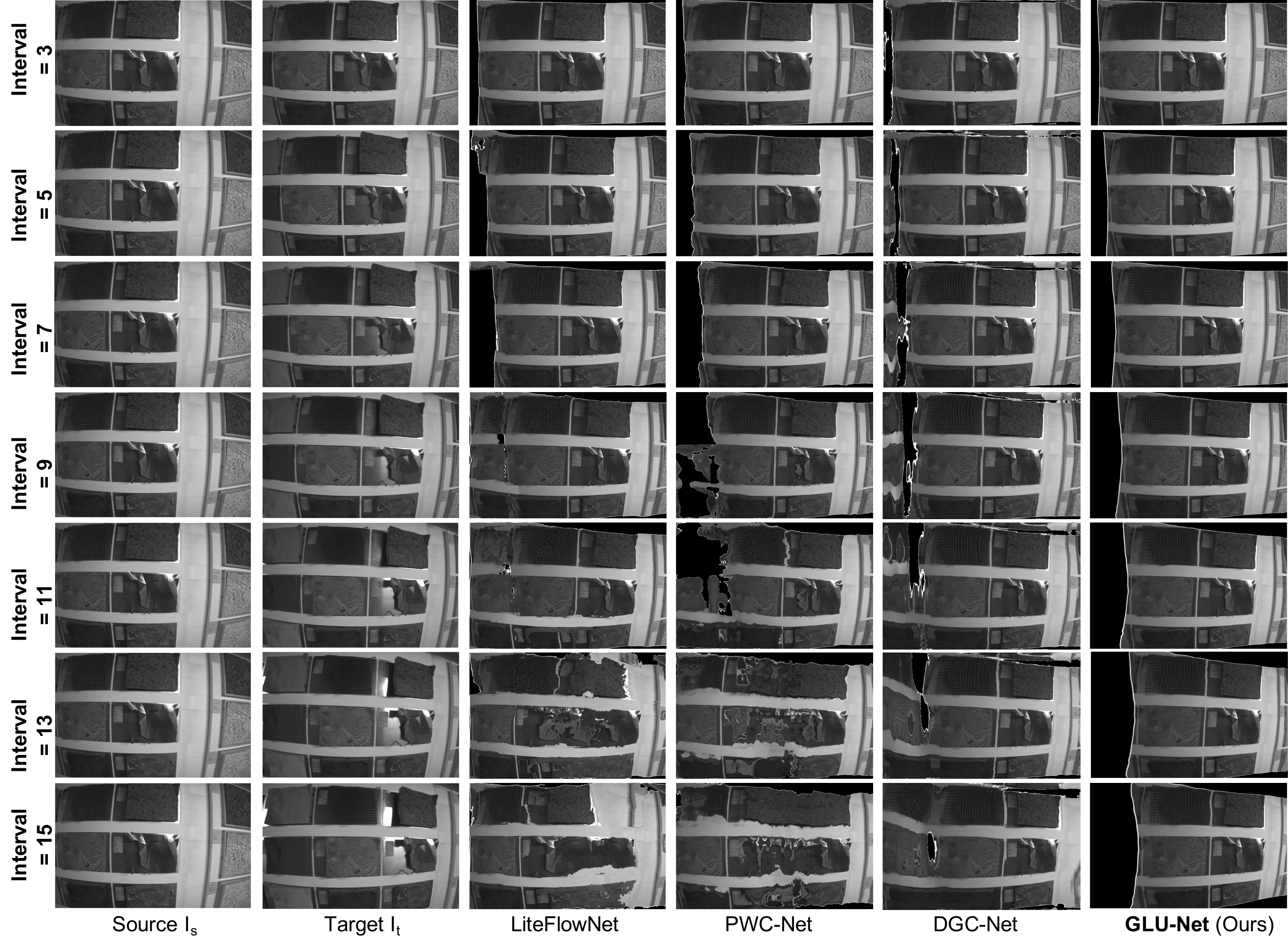}
\caption{Qualitative examples of multiple networks and our GLU-Net applied to pairs of ETH3D dataset \emph{taken at different intervals by the same camera}. The source images are warped according to the flow fields outputted by the different networks. The warped source images should resemble the target images. Optical flow methods obtain good qualitative results for low intervals (3 and 5) but largely degrade on bigger intervals. On the contrary, GLU-Net has a steady performance over all intervals. }
\label{ETH3d-intervals}
\end{figure*}

Here, we additionally provide qualitative examples of the different networks and GLU-Net applied to pairs of images at different intervals in Figure~\ref{ETH3d-intervals}. It is visible that while optical flow methods achieve good results for low intervals, the warped source images according to their outputted flows get worst when increasing the intervals between image pairs. On the other hand, our model produces flow fields of constant quality. 

\begin{figure*}[t]
\centering
\includegraphics[width=0.99\textwidth]{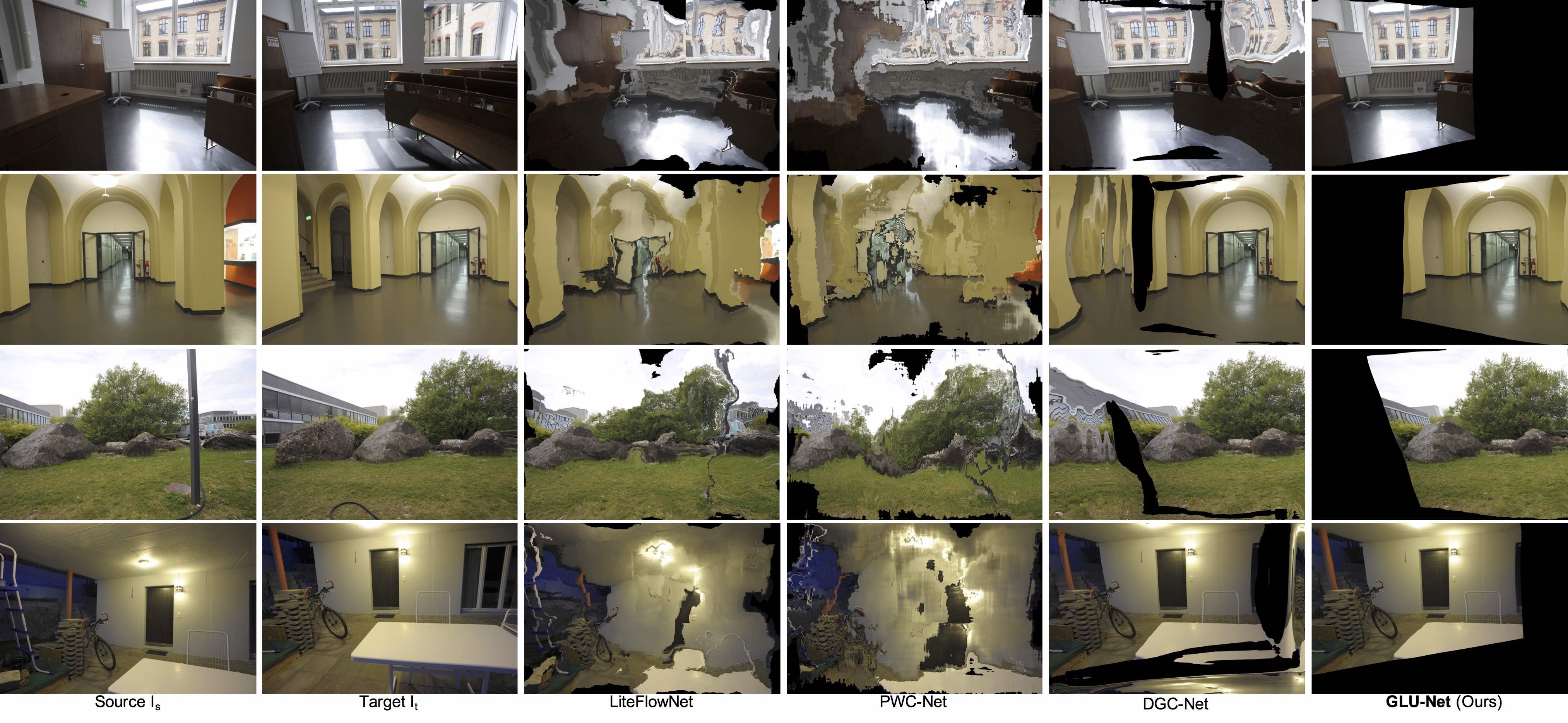}
\vspace{-2mm}\caption{Qualitative examples of state-of-the-art methods applied to very high-resolution images of different scenes of ETH3D. The presented image pairs show substantial view-point changes, and thus \emph{very large motions}.}
\label{ETH3d-hr}
\end{figure*}

\parsection{Qualitative results} We additionally use ETH3D images to demonstrate the superiority of our approach to deal with extreme viewpoint changes on the one hand, and radical illumination and appearance variations on the other hand. 

In addition to the medium resolution images evaluated previously, ETH3D~\cite{ETH3d} also provides several additional scenes taken with high-resolution cameras, acquiring images at 24 Megapixel ($6048 \times 4032$). Since the images of a sequence are taken by a unique camera, consecutive pairs of images show only little lightning variations, however they are related by \emph{very wide view-point changes}. 
As there are no ground-truth correspondences provided along with the images, we only evaluate qualitatively on consecutive pairs of images. The original images of $6048 \times 4032$ are down-samled by a factor of $2$ for practical purposes. We show quantitative results over a few of those images in Figure~\ref{ETH3d-hr}. GLU-Net is capable of handling very large motions, where the other methods partly (DGC-Net) or completely fail (PWC-Net and LiteFlowNet).

\begin{figure*}
\centering
\includegraphics[width=0.99\textwidth]{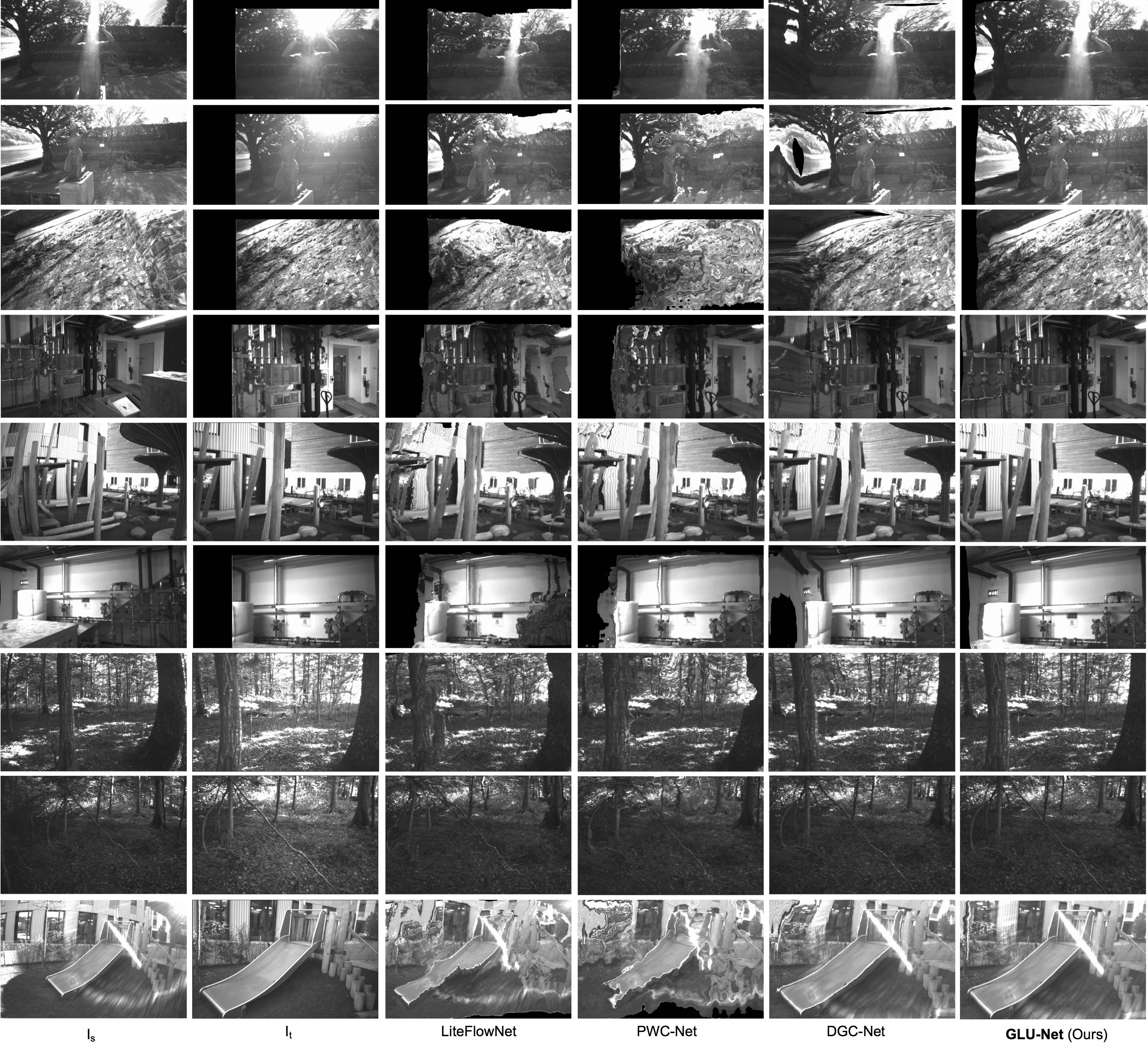}
\caption{Qualitative examples of ETH3D pairs of images taken \emph{simultaneously by two different cameras}. The two cameras have different field-of-views and sometimes different resolutions. Pairs of images experience drastic differences in lightning conditions.  The source images are warped according to the flow fields outputted by different state-of-the-art networks and by our GLU-Net. The warped source images should resemble the target images.}
\label{ETH3d-more}
\end{figure*}

On the other hand, our network can also handle large appearances changes due to variation in illumination or due to the use of different optics. For this purpose, we utilize additional examples of pairs of images from ETH3D taken by \emph{two different cameras} simultaneously. The camera of the first images has a field-of-view of 54 degrees while the other camera has a field of view of 83 degrees. They capture images at a resolution of $480 \times 752$ or $514 \times 955$ depending on the scenes and on the camera. The exposure settings of the cameras are set to automatic for all datasets, allowing the device to adapt to illumination changes. Qualitative examples of state-of-the-art methods and GLU-Net applied to such pairs of images are presented in Figure~\ref{ETH3d-more}. GLU-Net is robust to changes in lightning conditions as well as to artifacts. While DGC-Net~\cite{Melekhov2019} obtains satisfactory results, the warped image according to its outputted flow is often blurry whereas we always obtain sharp, almost perfect warped source images.

\clearpage
\newpage

\clearpage
\newpage
\begin{figure*}[t]
\vspace{-4mm}\centering
(A) Metrics with respect to rotation \\
\includegraphics[width=0.33\textwidth]{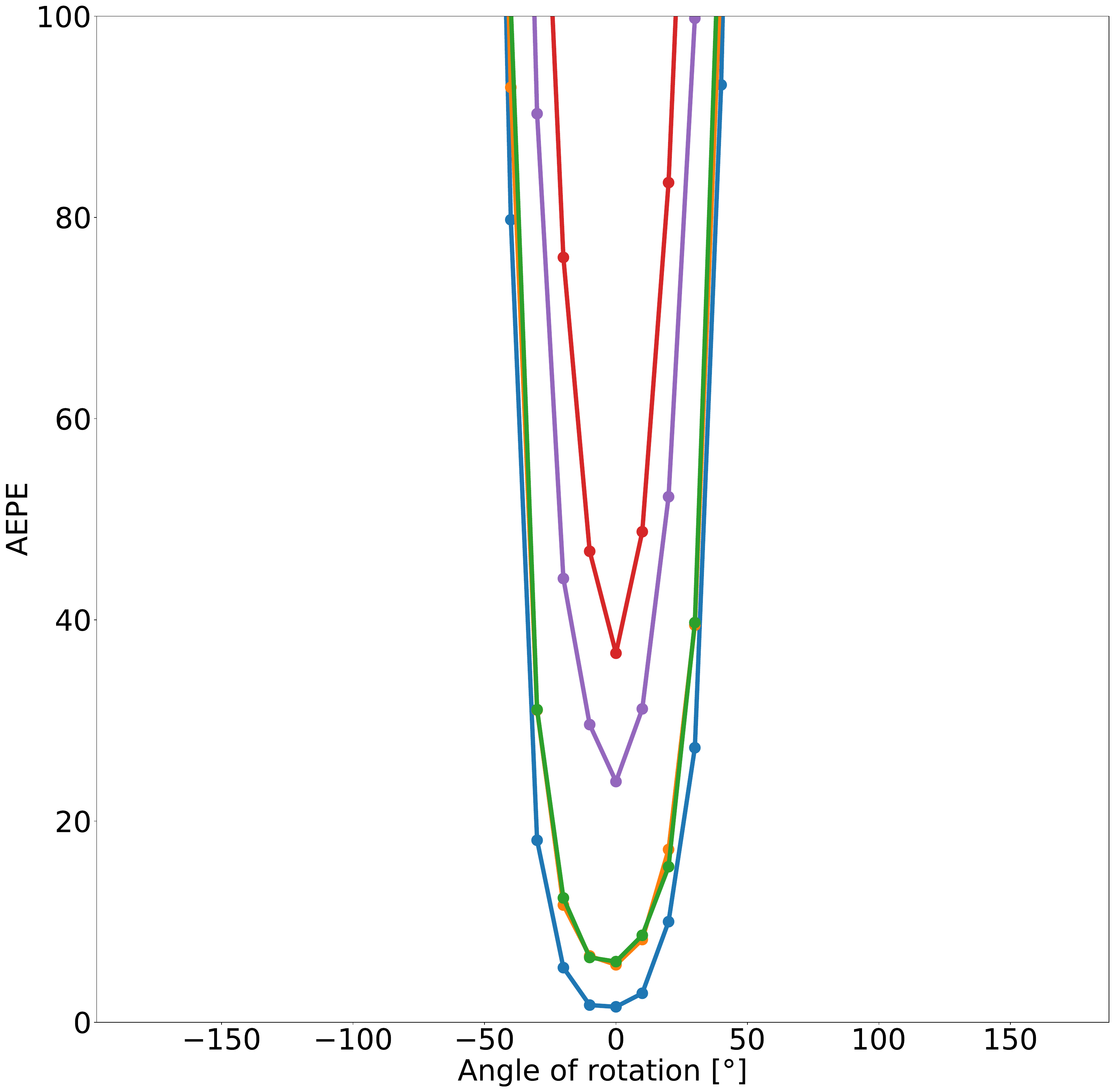}\hfill
\includegraphics[width=0.33\textwidth]{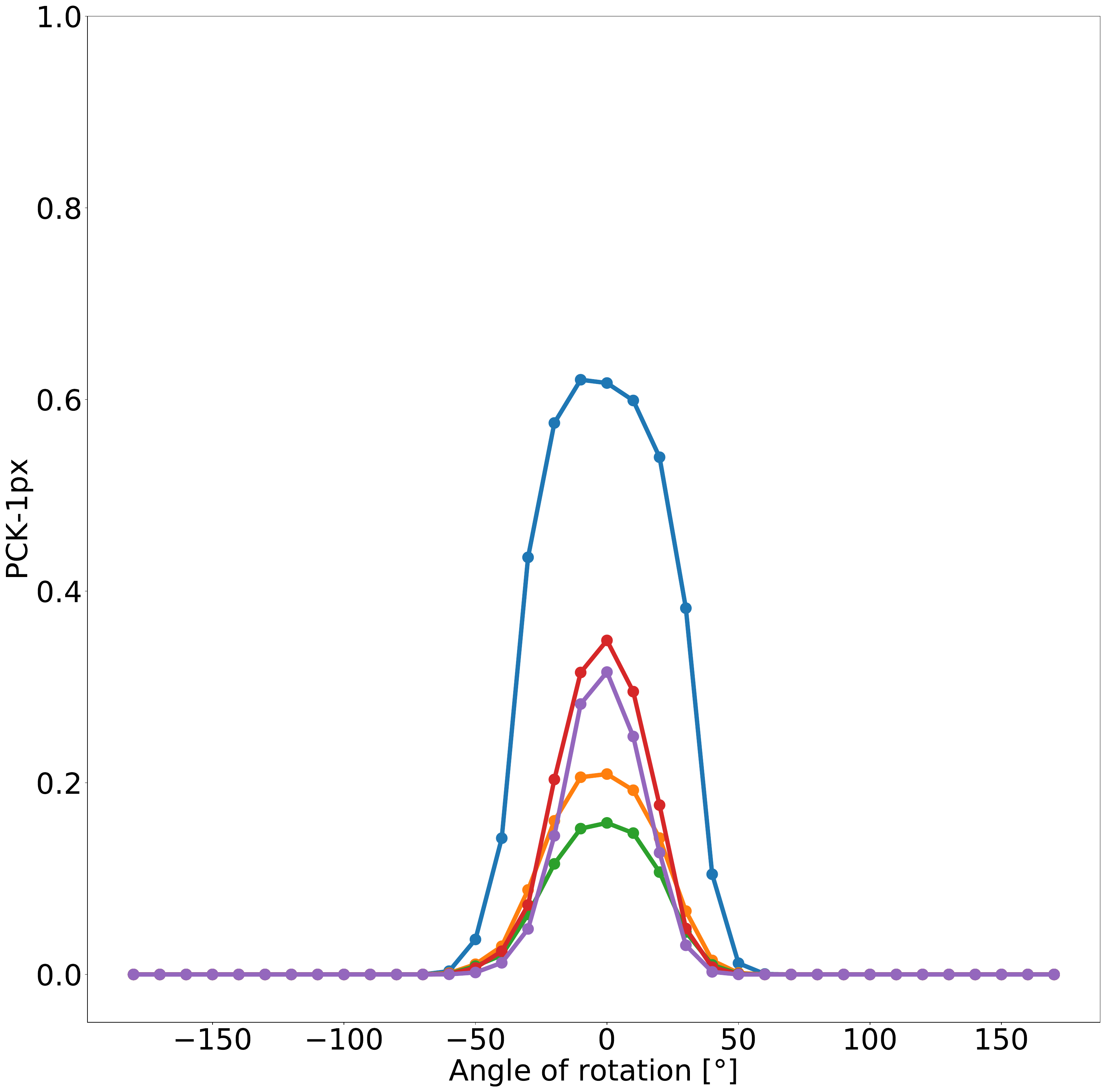}\hfill
\includegraphics[width=0.33\textwidth]{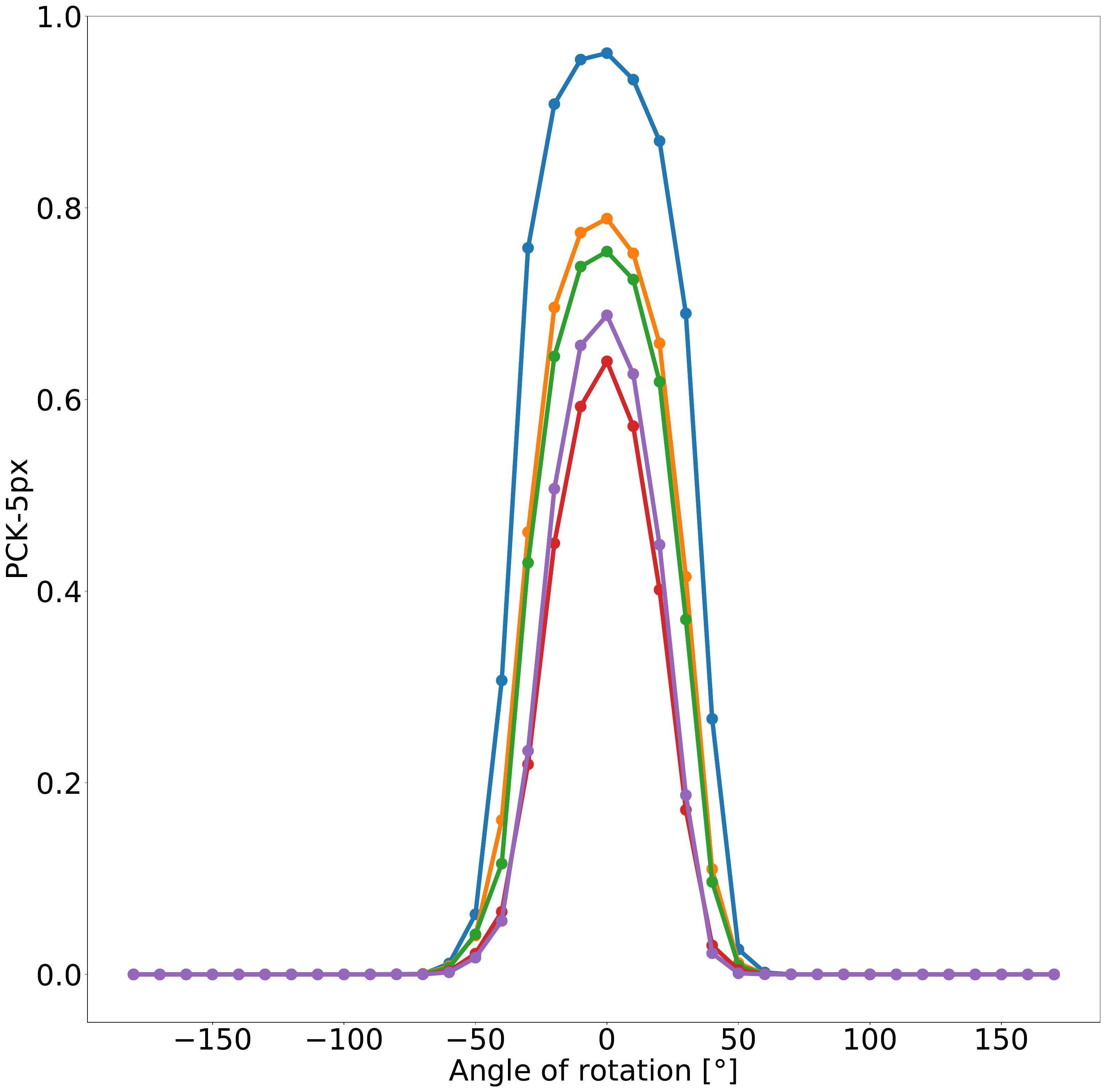}\hfill \\

(B) Metrics with respect to scaling \\
\includegraphics[width=0.33\textwidth]{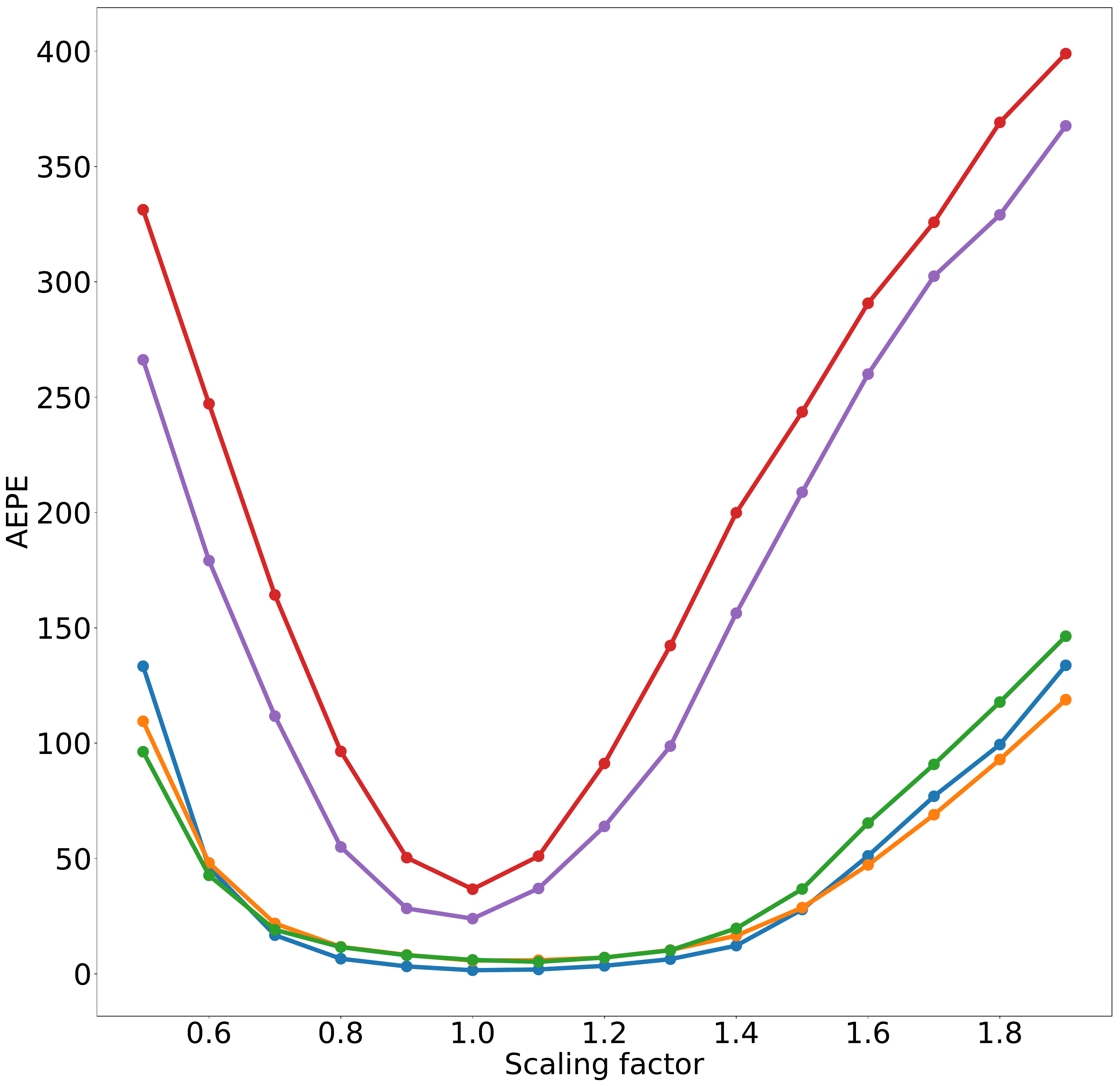}\hfill
\includegraphics[width=0.33\textwidth]{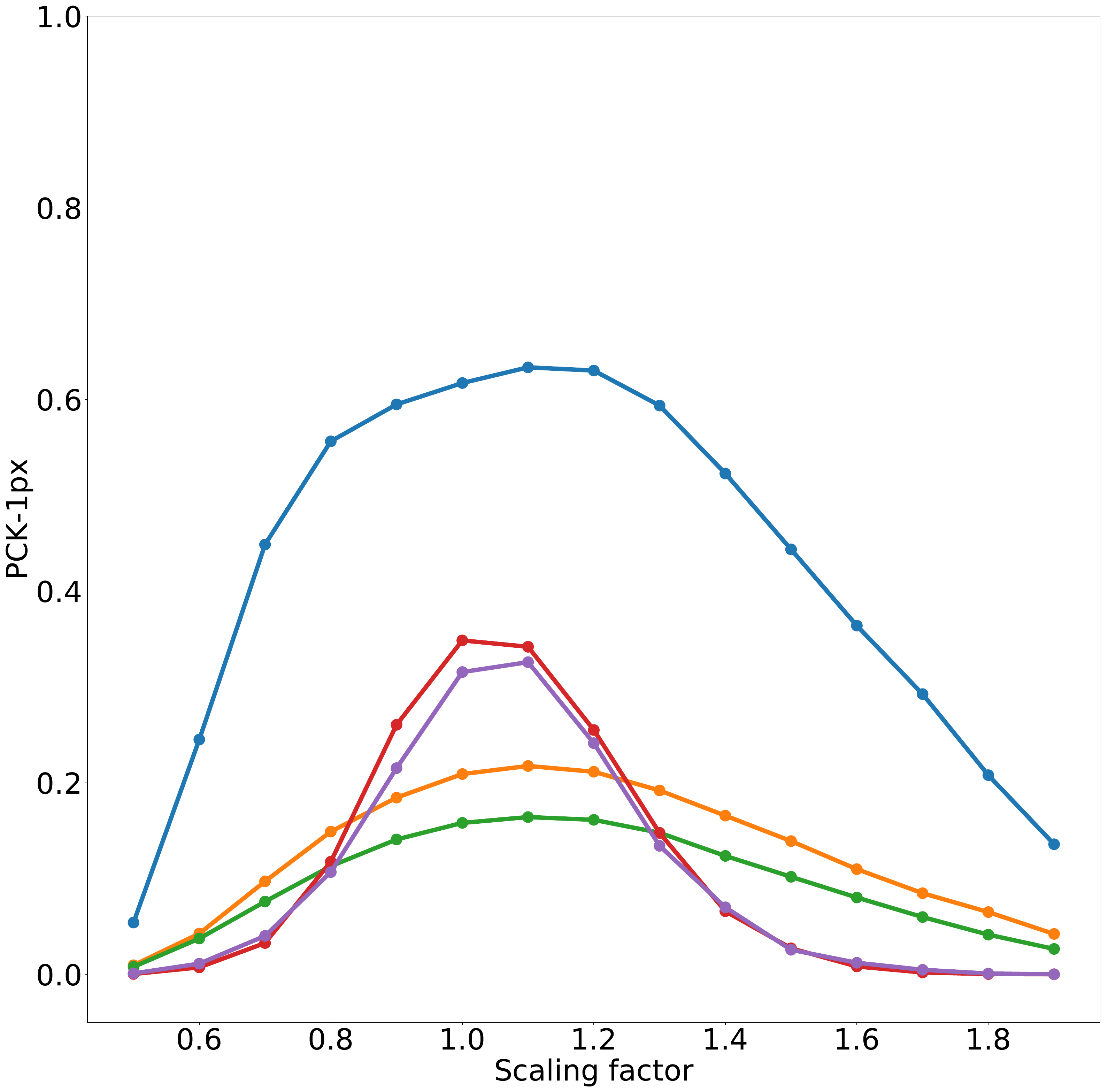}\hfill
\includegraphics[width=0.33\textwidth]{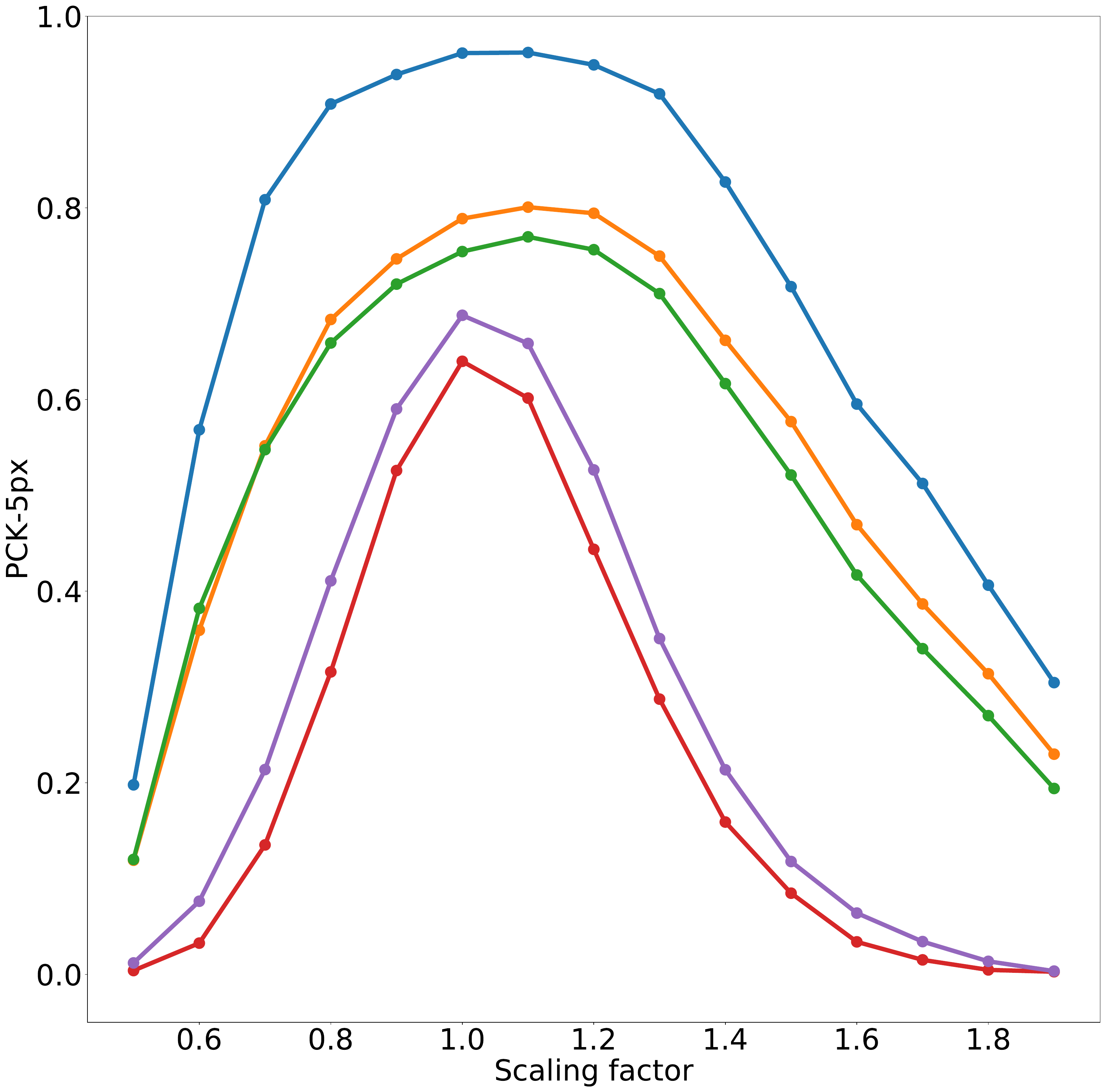}\hfill 
\includegraphics[width=0.98\textwidth]{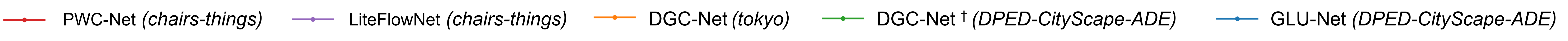}
\caption{Quantitative results (AEPE, PCK-1px and PCK-5px) over the first viewpoint of HP, for different rotation and scaling factors applied to the target images and ground-truth flow fields. The training datasets are indicated in parenthesis for each model.}
\label{rotation_sc_large}
\end{figure*}

\subsubsection{Rotation and scaling}

We additionally measured the performance of our GLU-Net compared to state-of-the-art networks with respect to increasing rotation and scaling factors. To do so, we used the 59 pairs of the ViewPoint I of the HP~\cite{Lenc} dataset as base images and applied increasingly high rotation and scaling factors to the target and ground-truth flow fields. In Figure~\ref{rotation_sc_large}, we plot the metrics (AEPE, PCK-1px and PCK-5px) obtained by GLU-Net, DGC-Net, DGC-Net$^\dagger$, PWC-Net and LiteFlowNet with respect to increasing applied rotation or scaling factors. 
For both rotation and scaling, while GLU-Net obtains similar AEPE than DGC-Net, its accuracy (PCK-1px and PCK-5px) is significantly above that of DGC-Net. 

It must also be noted that GLU-Net is particularly robust and accurate for rotations up to +/- 50 degrees and scaling factors comprised between 0.8 and 1.4. This corresponds to the extent of geometric transformations present in the training dataset. Therefore, for improved robustness to larger rotations or scaling, image pairs experiencing such transformations should be additionally included in the training set. 

\subsection{Semantic correspondences}
\label{semantic}

\begin{figure*}[t]
\centering
\includegraphics[width=0.99\textwidth]{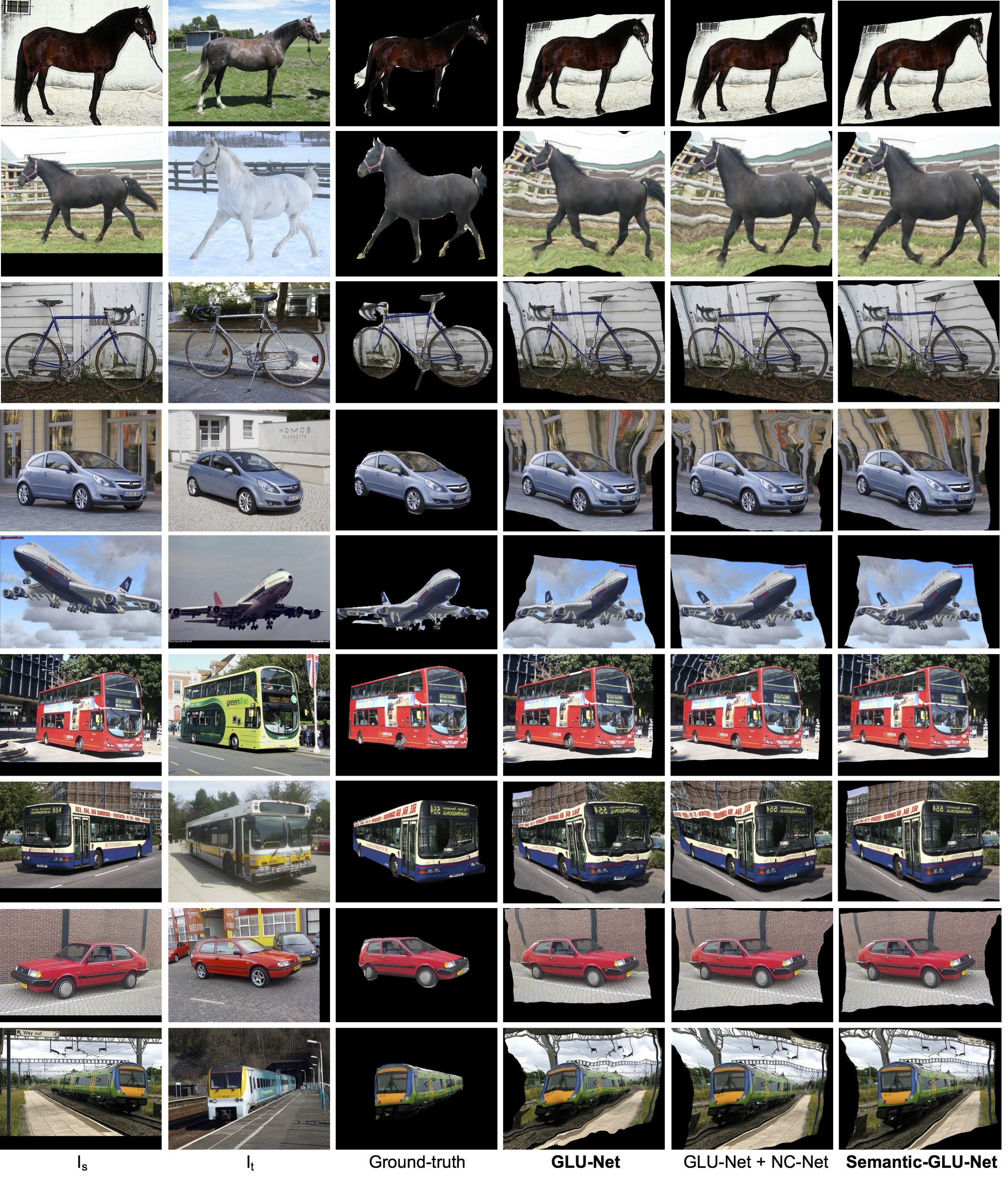}
\caption{Qualitative examples of our universal network GLU-Net as well as GLU-Net with specific architectural details from the semantic correspondence literature applied to TSS images. The additional architectural modules are the Neighborhood Consensus Network NC-Net~\cite{Rocco2018b} and concatenating features within the L-Net~\cite{Jeon}. Adopting those two modules leads to Semantic-GLU-Net. The source images are warped according to the flow fields outputted by the different networks. The warped source images should resemble the target images and the ground-truths.}
\label{TSS}
\end{figure*}

In Figure~\ref{TSS}, we present additional qualitative results on the TSS~\cite{Taniai2016} dataset of our universal network (GLU-Net) and its modified version (Semantic-GLU-Net), which includes NC-Net~\cite{Rocco2018b} and feature concatenation~\cite{Jeon}.

\begin{figure*}[t]
\centering
(A) KITTI-2012 dataset \\ \includegraphics[width=0.99\textwidth]{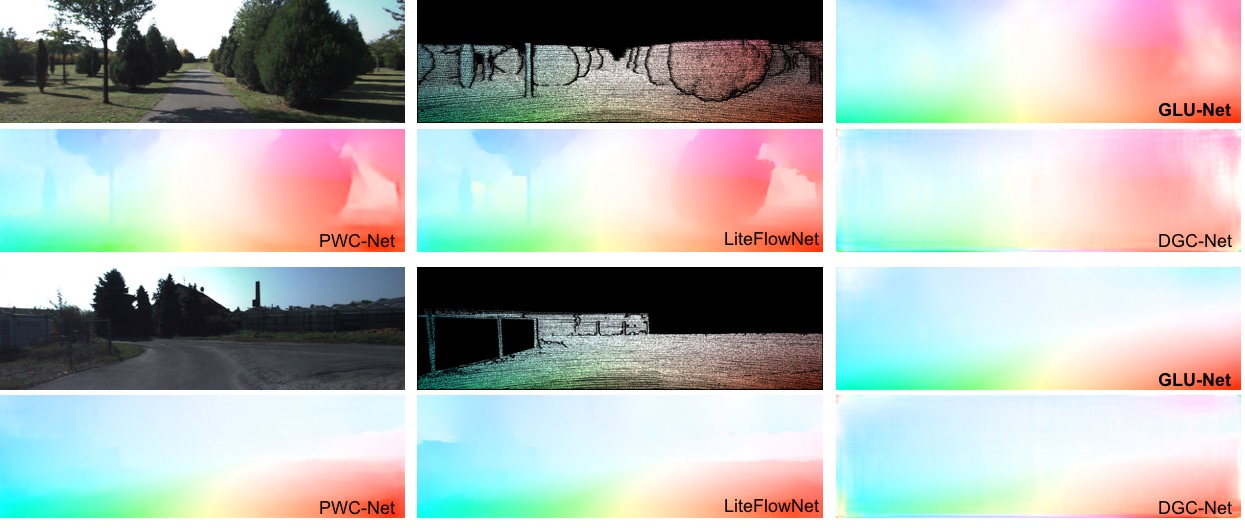} \\
(B) KITTI-2015 dataset \\
\includegraphics[width=0.99\textwidth]{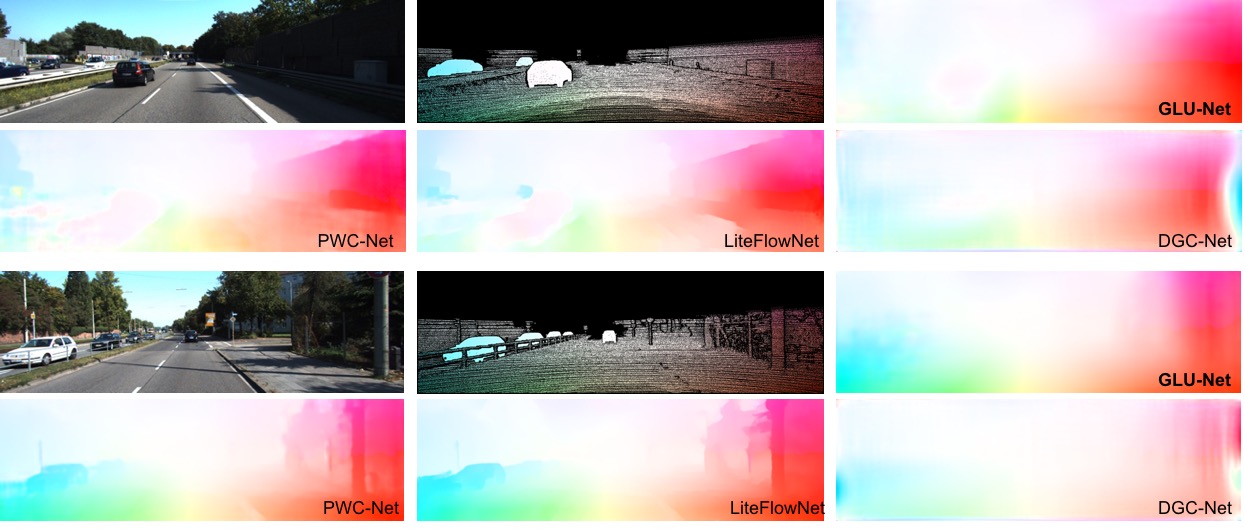}
\vspace{-1mm}\caption{Representation of the flow fields estimated by state-of-the-art methods and GLU-Net applied to images of : (A) KITTI 2012 dataset, that is restricted to static scenes; (B) KITTI 2015 dataset, which comprises dynamic scenes.}
\label{kitti-supp}
\end{figure*}

\subsection{Optical flow}
\label{OF}

\parsection{Additional qualitative results} In Figure~\ref{kitti-supp}, we present additional qualitative examples of the estimated flow fields obtained by our method and competitors on the KITTI datasets. While our approach GLU-Net lacks accuracy at the object boundaries compared to the optical flow methods, our results are substantially better than those of DGC-Net, which is trained on the same kind of synthetic geometric transformations. As already stated, improved results, particularly at the object boundaries, could be obtained by including optical flow data with independently moving objects in the training set. 

\begin{figure}[H]
\centering
\includegraphics[width=0.49\textwidth]{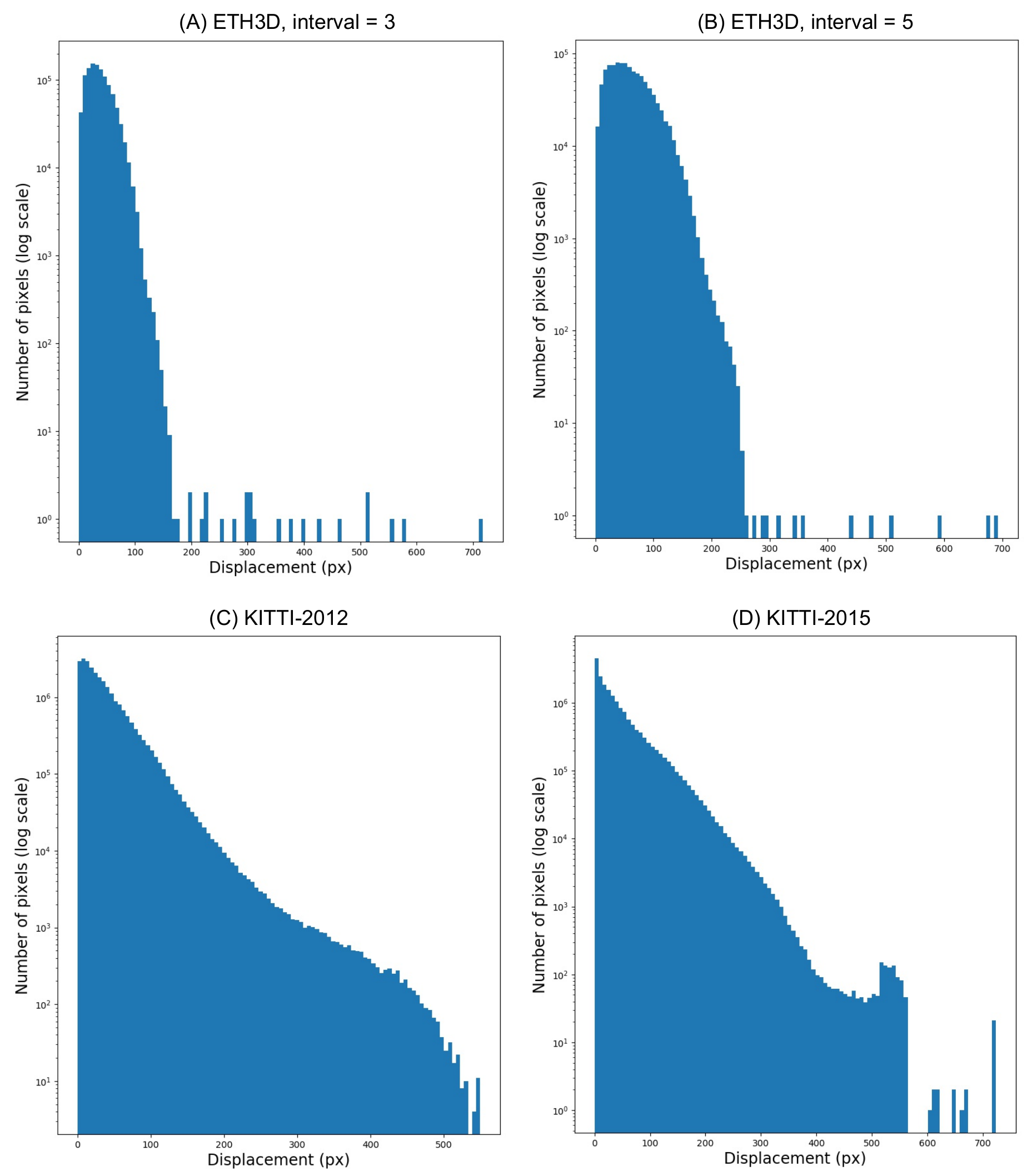} \\
\vspace{-2mm}\caption{Ground-truth flow distribution (in log scale) for the ETH3D dataset sampled at small intervals and for the KITTI datasets.}
\label{flow_distrib}
\end{figure}

\parsection{Supplementary analysis of the optical flow results}
According to Table~\ref{tab:KITTI} and Figure~\ref{ETH3d} of the main paper, our GLU-Net obtains better AEPE than the optical flow methods PWC-Net and LiteFlowNet on the KITTI datasets (Table~\ref{tab:KITTI}), but worst AEPE on the first intervals of ETH3D (Figure~~\ref{ETH3d}). The reasons for this behavior are explained below.
As observed in Figure~\ref{flow_distrib}, while KITTI and ETH3D pairs for small intervals show similar average displacement, the KITTI datasets have a much wider distribution of displacements due to moving objects and the fast camera forward motion. Besides, our GLU-Net performs substantially better on the large-displacement pixels of KITTI compared to 

\noindent
PWC-Net and LiteFlowNet, as evidenced in Table~\ref{tab:disp}. This explains the \emph{on-average} advantage of our approach (better AEPE), despite being slightly weaker for small displacements.

\begin{table}[t]
\centering
\resizebox{0.48\textwidth}{!}{%
\begin{tabular}{l|ccc|ccc}
\toprule
            & \multicolumn{3}{c}{\textbf{KITTI-2012}} & \multicolumn{3}{c}{\textbf{KITTI-2015}} \\
            & Small     & Medium   & Large   & Small    & Medium    & Large   \\ \midrule 
PWC-Net     &    0.63       &   1.58       &    10.36     &  0.94        &    2.89       &    28.65     \\
LiteFlowNet & \textbf{0.46}      & \textbf{1.24}     & 10.83   & \textbf{0.68}    & \textbf{2.32}      & 29.93   \\
DGC-Net     & 1.53      &  3.10       & 21.90   & 3.44     & 6.48      & 36.07   \\
GLU-Net     & 0.83    & 1.63     & \textbf{7.68}    & 2.25     & 4.87      & \textbf{23.01}  \\  \bottomrule
\end{tabular}%
}\vspace{1mm}
\caption{AEPE for different ground truth pixel-displacement categories on the KITTI datasets. Small is defined as $\left \| \mathbf{w}_{GT} \right \|_2 < 10$, Medium as  $10 \leq \left \| \mathbf{w}_{GT} \right \|_2 < 40$ and Large as $ 40 \leq \left \| \mathbf{w}_{GT} \right \|_2$. The EPE is averaged over all pixels of the dataset. }
\label{tab:disp}
\end{table}

\subsection{Detailed ablative analysis}
\label{ablation-study}

In this section, we provide additional ablation experiments. All networks are trained on \textit{CityScape-DPED-ADE} dataset.

\parsection{Coarse-to-fine-approach} We first defend the use of a coarse-to-fine approach with a feature pyramid. We report AEPE and PCK metrics for the flow estimates obtained at different levels of the feature pyramid of GLU-Net model in Table~\ref{tab:levels}.  On the flow field estimated at each level, we apply bilinear interpolation to the original image resolution and multiply the estimated flow with the corresponding scale factor for the levels of L-Net.  The end-point error decreases from the coarsest level to the highest level of the pyramid while the accuracy (PCK) increases. This supports the use of a pyramidal model. 

\parsection{Scale pyramid level of the adaptive resolution} In Table~\ref{tab:ab-level}, we present the influence of the pyramid level at which the adaptive resolution module is integrated in the four-level pyramid network. Having a single level in L-Net (corresponding to the global correlation layer) and three pyramid levels in H-Net (referred to as \textbf{3L}) lead to poor results on all datasets, even compared to GLOCAL-Net. On the other hand, both other alternatives (1 or 2 levels in H-Net) bring about major improvements of robustness (AEPE) and accuracy (PCK) on HPatches dataset, particularly on the high-resolution images HP. However, having only one level in H-Net (\textbf{1L}) degrades the performances obtained on the semantic dataset TSS. H-Net and L-Net both comprised of 2 pyramid levels (\textbf{2L}) appears as the best option to achieve competitive results on geometric matching, optical flow as well as semantic matching. 

\begin{table}
\centering
\resizebox{0.48\textwidth}{!}{%
\begin{tabular}{lllllll}
\toprule
& AEPE & PCK-1px [\%] & PCK-5px [\%]\\\midrule
Level 1 [$16 \times 16$] &  45.49 & 0.70 & 13.53\\ 
Level 2 [$32 \times 32$] &   30.00  & 6.27 &  50.29\\ 
Level 3 [$H/8 \times W/8$] &   26.43 & 30.47  &  74.44\\ 
Level 4 [$H/4 \times W/4$] & 25.05  & 39.55  & 78.54  \\ \bottomrule
\end{tabular}%
}    \vspace{1mm}
\caption{Effect of coarse-to-fine approach for our GLU-Net: Metrics calculated over HP images. The flow estimated at each pyramid level is up-sampled to original image resolution and the metrics are calculated at this resolution.}
\label{tab:levels}
\end{table}

\begin{table}
\centering
\resizebox{0.49\textwidth}{!}{%
\begin{tabular}{llcccc}
\toprule
&& GLOCAL-Net & 1L = 1 H-Net level & 2L = 2 H-Net levels & 3L = 3 H-Net levels \\ \midrule

\multirow{3}{6mm}{\textbf{HP-240}}
&AEPE &  8.77 & \textbf{7.47} & 7.69 & 8.93 \\
&PCK-1px [\%] & 48.53 & \textbf{62.85} & 53.83 & 35.81 \\
&PCK-5px [\%] & 78.12 & \textbf{85.32} & 83.17 & 75.97 \\ \midrule

            &AEPE & 31.64 & \textbf{24.75} & 25.55 & 32.03 \\
\textbf{HP} &PCK-1px [\%] & 10.23 & 33.92 & \textbf{35.26} & 28.76 \\
            &PCK-5px [\%] & 56.73 & \textbf{76.99} & 75.79   & 69.78\\ \bottomrule
\textbf{TSS}
&PCK [\%]    &  77.29  &  62.98  &      \textbf{78.97}        & 69.78\\ \bottomrule
\end{tabular}%
}\vspace{1mm}
\caption{Effect of adaptive resolution and its position. All networks are without iterative refinement and without cyclic consistency. 2 H-Net levels (\textbf{2L}) is the only alternative for a universal network applicable to geometric matching, semantic correspondence and optical flow. }\label{tab:ab-level}
\end{table}

\end{document}